\documentclass[runningheads]{llncs}

\usepackage{eccv}

\usepackage{eccvabbrv}

\usepackage{graphicx}
\usepackage{booktabs}

\usepackage{multirow}
\usepackage[table]{xcolor}
\usepackage{colortbl}
\usepackage{caption}
\usepackage{amssymb}

\usepackage[accsupp]{axessibility}  

\usepackage{hyperref}

\usepackage{orcidlink}

\begin{document}

\title{SPLIT: Training-Free AI-Generated and Partially Edited Video Detection via Spatial Patch‑Level Incoherence and Temporal Roughness} 

\newcommand{\framework}{\textsc{SPLIT}}
\newcommand{\dash}{\textemdash\allowbreak}

\titlerunning{SPLIT: Training-Free AI-Generated and Edited Video Detection}

\author{
Jongyeop Hyun\inst{1}\orcidlink{0009-0006-3911-4015}
\and
Hyounghun Kim\inst{1,2}\thanks{Corresponding author.}\orcidlink{0009-0008-3382-7510}
}

\authorrunning{J.~Hyun and H.~Kim}

\institute{
Graduate School of Artificial Intelligence, POSTECH
\and
Department of Computer Science and Engineering, POSTECH\\
\email{\{mldljyh,h.kim\}@postech.ac.kr}
}

\maketitle

\begin{abstract}
Deploying AI-generated video detectors in real-world services demands an ultra-low false positive rate (FPR) on real videos to avoid falsely rejecting authentic content, a regime where standard metrics such as AUROC fail to reflect actual operating behavior. We introduce \textbf{S}patial \textbf{P}atch‑\textbf{L}evel \textbf{I}ncoherence and \textbf{T}emporal Roughness (\framework{}), a training-free detector that operates on patch tokens from a frozen vision encoder to detect both fully generated and partially edited videos. \framework{} computes two complementary signals: Two-step Temporal Roughness (TTR), capturing non-smooth patch trajectories via one-step and two-step feature variation contrast, and Local Spatial Motion Incoherence (LSMI), measuring spatially inconsistent temporal changes through gradients of a feature-space motion field. The two are fused multiplicatively with gamma correction to sharpen real--fake separation at strict thresholds. We further propose a service-aligned evaluation protocol based on Fake Recall at fixed FPR with real-only threshold calibration and cross-real threshold transfer. Across three benchmarks---FakeParts, GenVideo, and ViF-Bench---\framework{} achieves the highest Fake Recall at FPR = $0.1\%$, substantially outperforming supervised and training-free baselines while remaining robust to post-processing with negligible overhead. The code is publicly available at \url{https://github.com/mldljyh/SPLIT}.
\keywords{AI-generated video detection \and Training-free detection \and Partial video manipulation}
\end{abstract}

\section{Introduction}
\label{sec:introduction}
\begin{figure}[t]
  \centering
  \includegraphics[width=\textwidth]{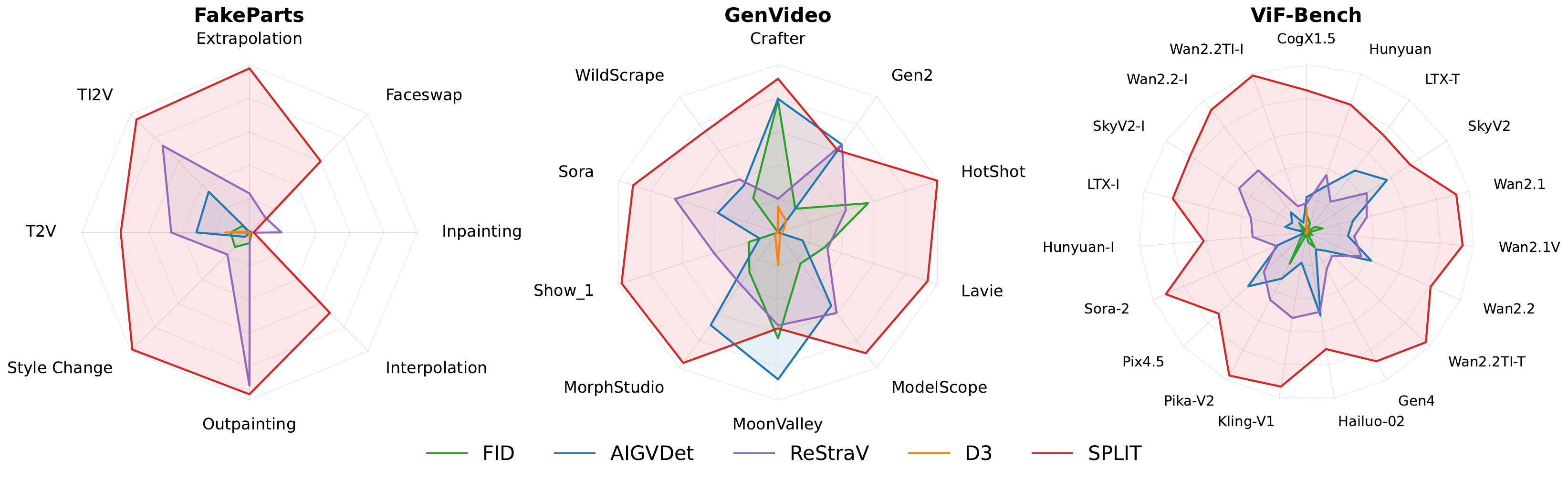}
  \caption{Fake Recall (\%) at FPR = 0.1\% for FID, AIGVDet,  ReStraV, D3, and \framework{} across FakeParts, GenVideo, and ViF-Bench. \framework{} (red) consistently achieves the highest and most uniform recall across all manipulation types and generators.}
  \label{fig:teaser_01}
\end{figure}

Recent progress in text-to-video and image-to-video generation has made high-quality synthetic video broadly accessible, while also enabling subtle \emph{partial} edits such as localized inpainting/outpainting, temporal interpolation/extrapolation, faceswap, and style transfer~\cite{blattmann2023stable,chen2024videocrafter2,openai2025sora,wan2025wana,klinga}. This creates an urgent need for reliable AI-generated video detection in deployment settings with stringent operating constraints: false rejections of authentic user content must remain extremely rare (\eg, FPR on real videos~$\le 0.1\%$), yet recall on manipulated content must stay high~\cite{Xiong_2026_WACV}. In this regime, aggregate metrics such as AUROC can be misleading because they do not directly reflect behavior at ultra-low FPR~\cite{Xiong_2026_WACV, hand2009measuring}.

Prior work has largely focused on supervised detectors that exploit spatial artifacts, frequency cues, or spatiotemporal inconsistencies, often leveraging large pretrained video encoders~\cite{guera2018deepfake,li2019exposing,zheng2021exploring,xu2023tall,ni2022expanding,chen2024demamba,tan2024rethinking,gu2021spatiotemporala,bai2025aigenerated}. While effective within their training distributions, these methods can degrade under distribution shift from new generators, post-processing, and especially partial edits where most of the content remains real~\cite{liu2025fakeparts}. Training-free approaches such as D3~\cite{zheng2025d3a} improve cross-generator generalization via clip-level second-order temporal volatility in frozen features, but pooling over the clip can dilute sparse, localized manipulation evidence---as highlighted by partial-edit benchmarks such as FakeParts~\cite{liu2025fakeparts}.

This paper targets the intersection of three practical requirements: (i)~detecting both fully generated and partially edited videos, (ii)~generalizing across generators without task-specific training, and (iii)~operating under service-aligned ultra-low-FPR constraints with thresholds calibrated using real data only. We introduce \textbf{S}patial \textbf{P}atch-\textbf{L}evel \textbf{I}ncoherence and \textbf{T}emporal Roughness (\framework{}), a training-free detector that operates on \emph{patch tokens} from a frozen pretrained vision encoder. Rather than collapsing a frame into a single embedding, \framework{} preserves localized spatiotemporal evidence and aggregates it into a single manipulation score without any learned parameters.

\framework{} computes two complementary patch-token signals. \textbf{Two-step Temporal Roughness (TTR)} measures deviations from smooth temporal evolution by comparing accumulated one-step and two-step feature variations, capturing artifacts such as flicker and local temporal misalignment. \textbf{Local Spatial Motion Incoherence (LSMI)} measures whether temporal changes are spatially coordinated by constructing a feature-space motion field and computing local spatial gradients; manipulated regions tend to induce spatially inconsistent temporal changes. We fuse these signals multiplicatively with a gamma correction to amplify subtle but consistent roughness differences.

We also advocate deployment-constrained evaluation, reporting \emph{Fake Recall at FPR} = $\alpha$ for $\alpha \in \{0.1\%, 1\%, 5\%\}$, with thresholds calibrated on real videos only, and introduce \emph{cross-real threshold transfer} to quantify calibration stability when the real-data source shifts. We evaluate on three complementary benchmarks: FakeParts~\cite{liu2025fakeparts} (partial edits), GenVideo~\cite{chen2024demamba} (earlier generators; near-domain for supervised baselines), and ViF-Bench~\cite{li2025skyra} (newer 2024--2025 generators; out-of-distribution stress test). As shown in \cref{fig:teaser_01} at FPR~$=$~$0.1\%$, \framework{} outperforms representative supervised baselines and the training-free D3 across manipulation types and generators. Our contributions are as follows:
\begin{itemize}
    \item We introduce \textbf{\framework{}}, a training-free patch-token detector that measures localized spatiotemporal incoherence through \textbf{TTR} and \textbf{LSMI}, fused with gamma correction to sharpen separation under strict thresholds.
    \item We formalize service-aligned AI-generated video detection via Fake Recall at ultra-low FPR with real-only threshold calibration, and propose cross-real threshold transfer to measure calibration stability across heterogeneous real domains.
    \item We demonstrate strong generalization and robustness across partial edits and diverse generators on FakeParts, GenVideo, and ViF-Bench, with negligible overhead beyond the frozen encoder forward pass.
\end{itemize}
\section{Related Work}
\label{sec:related_work}

\textbf{AI-Generated Video Detection.}
Early deepfake detection targeted face-centric manipulations using spatial artifacts~\cite{guera2018deepfake, li2019exposing}, later incorporating temporal modeling for cross-frame consistency~\cite{zheng2021exploring, xu2023tall}. With the rise of diffusion-based generators~\cite{blattmann2023stable, chen2024videocrafter2}, recent detectors target broader video-level artifacts beyond faces. These approaches span multiple strategies: frequency-aware methods like NPR~\cite{tan2024rethinking} exploit upsampling artifacts that persist across generators; spatiotemporal methods such as STIL~\cite{gu2021spatiotemporala} learn joint appearance-dynamics inconsistencies, while FTCN~\cite{zheng2021exploring} and MINTIME~\cite{coccomini2024mintime} model temporal coherence primarily for face forgery scenarios. Foundation-model-based approaches leverage pretrained encoders---XCLIP~\cite{ni2022expanding} extends CLIP~\cite{radford2021learning} with temporal attention, and DeMamba~\cite{chen2024demamba} augments such encoders with a Mamba~\cite{gu2024mamba} module that captures local spatiotemporal inconsistencies through continuous scanning of partitioned regions. AIGVDet~\cite{bai2025aigenerated} combines spatial CNN features with optical-flow branches for video-level fusion.

A complementary frozen-feature line uses representation dynamics rather than training large video detectors. ReStraV~\cite{NEURIPS2025_1d9a4375}, inspired by perceptual straightening, observes that natural videos trace straighter trajectories than AI-generated videos in DINOv2 feature space, summarizing frame-level step distances and curvature statistics to train a lightweight classifier. D3~\cite{zheng2025d3a} is fully training-free and measures clip-level second-order temporal volatility via central differences in frozen features. These methods show that frozen representation dynamics provide useful cross-generator cues, but they rely on global frame/clip summaries that can dilute sparse manipulation evidence. In contrast, \framework{} operates on patch tokens and explicitly targets localized spatiotemporal incoherence: TTR compares accumulated 1-step and 2-step patch changes, while LSMI measures spatial disagreement of patch-wise temporal motion. This patch-level formulation is especially suited to partial edits, where manipulation evidence is region-specific.

\noindent
\textbf{Benchmarks and Partial Edits.} Several benchmarks have been developed for AI-generated video detection~\cite{chen2024demamba,liu2024evalcrafter,wang2024vidprom,ni2025genvidbencha,li2025skyra}. GenVideo~\cite{chen2024demamba} provides a million-scale dataset covering diverse generators with cross-generator and degraded-video evaluation tasks. ViF-Bench~\cite{li2025skyra} extends coverage to the latest generators including Sora-2~\cite{openai2025sora}, Wan2.2~\cite{wan2025wana}, and Kling~\cite{klinga}, with fine-grained artifact annotations. These benchmarks primarily target fully AI-generated videos (T2V, TI2V), driving progress on global detectors.
FakeParts~\cite{liu2025fakeparts} addresses a gap by introducing localized spatial (inpainting, outpainting, faceswap), temporal (interpolation, extrapolation), and style manipulations applied to otherwise real videos, with pixel- and frame-level annotations. Their evaluation reveals performance drops of up to 50\% for many detectors on partial edits, confirming that preserved real context defeats global approaches---a finding that highlights the value of our patch-level formulation~\cite{liu2025fakeparts}. While some recent deepfake benchmarks report strict operating points~\cite{Xiong_2026_WACV}, AI-generated video and partial-edit benchmarks still predominantly emphasize aggregate metrics such as AUROC, which can obscure behavior at ultra-low false-positive rates. To our knowledge, they do not evaluate real-only threshold calibration at FPR targets as low as $0.1\%$ or the stability of such thresholds under shifts in the real-data domain; we address this via Fake Recall at FPR $\in \{0.1\%, 1\%, 5\%\}$ with real-only calibration and cross-real threshold transfer.

\section{Method}
\label{sec:method}

\begin{figure}[t]
  \centering
  \includegraphics[width=\textwidth]{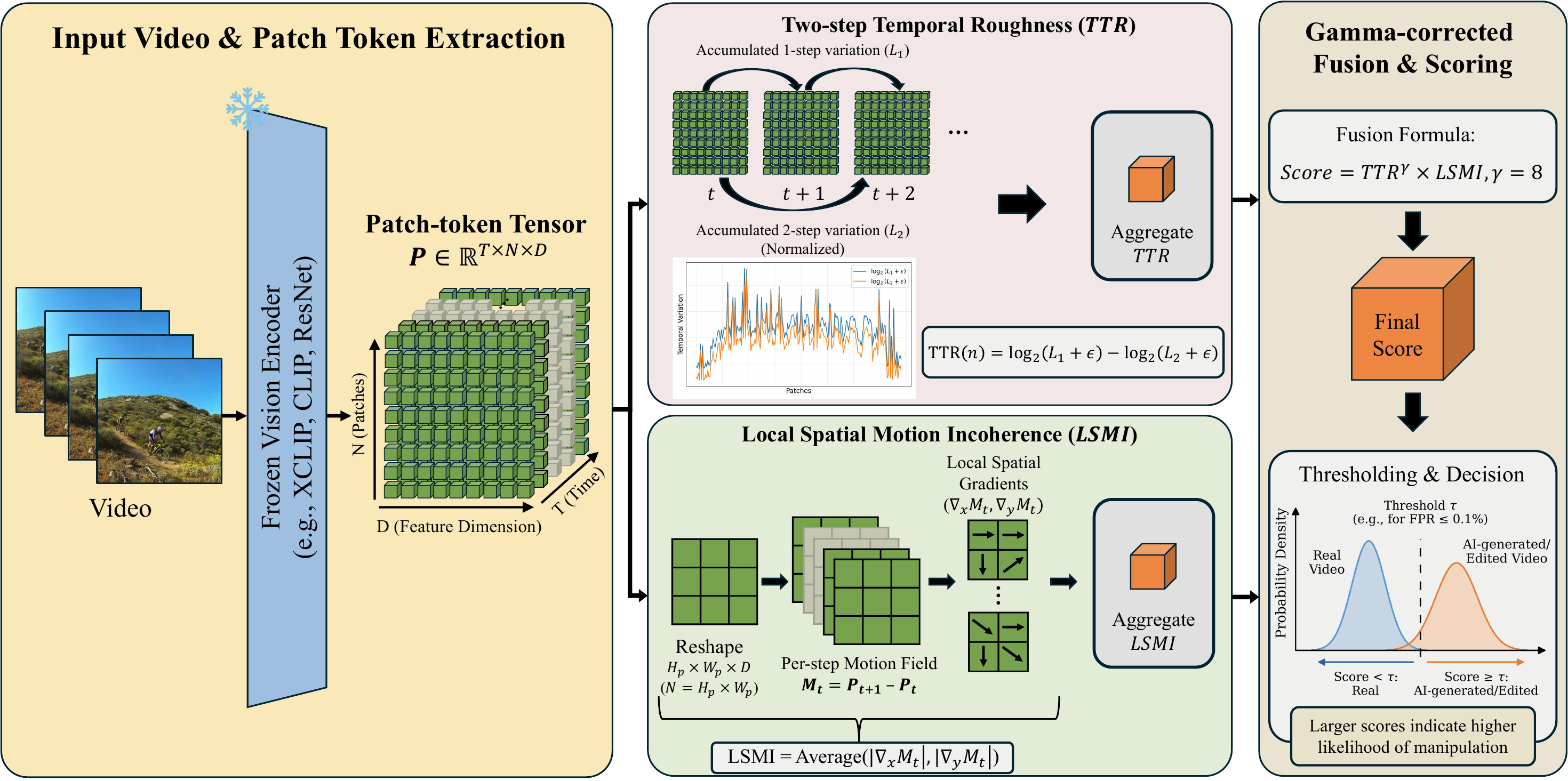}
   \caption{Overview of \framework{}. A frozen vision encoder produces patch tokens $P \in \mathbb{R}^{T \times N \times D}$, from which two signals are computed: \emph{Two-step Temporal Roughness (TTR)}, comparing 1-step and 2-step feature variations, and \emph{Local Spatial Motion Incoherence (LSMI)}, measuring spatial gradients of a feature-space motion field. These are combined via gamma-corrected multiplicative fusion and thresholded using real-only calibration to classify videos as real or AI-generated/edited.}
  \label{fig:method_overview}
\end{figure}

\subsection{Overview}

We propose Spatial Patch-Level Incoherence and Temporal Roughness (\framework{}), a training-free detector for AI-generated and partially edited videos under service-aligned operating constraints where the false positive rate (FPR) on real videos must remain extremely low (\eg, $\le 0.1\%$). \framework{} analyzes a clip using patch-level representations from a frozen pretrained vision encoder and computes two complementary signals: (1) Two-step Temporal Roughness (TTR), which measures how strongly patch features deviate from smooth temporal evolution by comparing 1-step and 2-step temporal variations; and (2) Local Spatial Motion Incoherence (LSMI), which measures spatial disagreement of patch-wise temporal changes by computing local spatial gradients of a feature-space motion field. These signals are fused multiplicatively with a gamma exponent to obtain a final score, where larger values indicate a higher likelihood of manipulation. An overview of \framework{} is shown in \cref{fig:method_overview}.

\subsection{Patch Token Extraction}

Given an input video clip $x \in \mathbb{R}^{T \times 3 \times H \times W}$, we extract per-frame patch tokens using a frozen pretrained vision encoder. Let $Z_t \in \mathbb{R}^{N \times D}$ denote the $N$ patch tokens of dimension $D$ at time $t$.

For transformer encoders (\eg, CLIP~\cite{radford2021learning}, XCLIP~\cite{ni2022expanding}, DINO~\cite{oquab2023dinov2}), we use the final-layer patch tokens and exclude the class token (\texttt{[CLS]}), yielding $Z_t \in \mathbb{R}^{N \times D}$. For CNN encoders (\eg, ResNet~\cite{he2016deep}, VGG~\cite{simonyan2015very}, EfficientNet~\cite{tan2019efficientnet}, MobileNet~\cite{howard2019mobilenetv3}), we extract the last convolutional feature map $F_t \in \mathbb{R}^{C \times H' \times W'}$ and treat each spatial cell as a token by reshaping to $Z_t \in \mathbb{R}^{(H'W') \times C}$, so that $N = H'W'$ and $D=C$. Stacking across time gives a patch-token tensor
\begin{equation}
    P = \{p_{t,n}\} \in \mathbb{R}^{T \times N \times D},
\end{equation}
where $p_{t,n} \in \mathbb{R}^D$ is the feature token of patch $n$ at time $t$. To support spatial neighborhood operations, we also view tokens on a 2D grid of size $H_p \times W_p$ such that $N = H_p W_p$. For ViT-style patchification, $(H_p, W_p)$ corresponds to the patch grid; for CNN feature maps, $(H_p, W_p) = (H', W')$.

\subsection{Two-step Temporal Roughness (TTR)}

\framework{} targets localized temporal artifacts that frequently arise from generation or partial edits, such as patch-wise flicker, inconsistent texture refresh, or local temporal misalignment. Rather than relying on a single notion of global volatility, TTR measures two-step consistency of temporal evolution in representation space by comparing accumulated 1-step and 2-step changes. For each patch $n$, we define the accumulated 1-step variation as
\begin{equation}
    L_1(n) = \sum_{t=1}^{T-1} \| p_{t+1,n} - p_{t,n} \|_2,
\end{equation}
and the accumulated 2-step variation $\hat{L}_2(n)$ together with its normalized form $L_2(n)$, which accounts for step length and short sequences, as
\begin{equation}
    \hat{L}_2(n) = \sum_{t=1}^{T-2} \| p_{t+2,n} - p_{t,n} \|_2,
    \qquad
    L_2(n) = \left(\frac{\hat{L}_2(n)}{2}\right)\cdot \frac{T-1}{T-2},
\end{equation}
which requires $T \ge 3$. TTR is then defined as a log-ratio:
\begin{equation}
    \mathrm{TTR}(n) = \log_2(L_1(n) + \epsilon) - \log_2(L_2(n) + \epsilon),
\end{equation}
where $\epsilon$ is a small constant for numerical stability. $L_1(n)$ is the consecutive-step path length of patch $n$'s feature trajectory, while $L_2(n)$ is a normalized two-step chord-length proxy. For constant-velocity evolution, the normalization gives $L_2(n)=L_1(n)$, and the two remain close for smooth motion. Irregular temporal evolution, such as flicker, texture refresh, or local temporal misalignment, makes successive patch-token displacements less aligned, increasing $L_1(n)$ relative to $L_2(n)$ and therefore increasing $\mathrm{TTR}(n)$. Thus, TTR is best interpreted as a statistical patch-level cue, not a necessary property of every fake clip: generated or edited videos are more likely to contain locally rough token trajectories, while real videos may also exhibit such behavior under cuts, shake, or abrupt motion. Our real-only threshold calibration accounts for this natural tail of the real-video score distribution. The log-ratio reduces sensitivity to absolute motion magnitude. We convert patch-wise roughness into a clip-level statistic by averaging in log-space:

\begin{equation}
    \mathbf{TTR} = \frac{1}{N}\sum_{n=1}^{N}\mathrm{TTR}(n).
\end{equation}
This choice has two motivations. First, since $\mathrm{TTR}(n)=\log_2\!\left(\frac{L_1(n)+\epsilon}{L_2(n)+\epsilon}\right)$, the average corresponds to the log of the geometric mean of per-patch roughness ratios,
$\mathbf{TTR}=\log_2\!\left(\prod_n \frac{L_1(n)+\epsilon}{L_2(n)+\epsilon}\right)^{1/N}$,
which aggregates multiplicative evidence while reducing sensitivity to extreme outliers. Second, mean pooling estimates the expected local temporal roughness across the frame and yields a low-variance score, which is important for our real-only threshold calibration at ultra-low FPR. This patch-level design preserves sensitivity to localized edits while remaining robust to content and object motion through aggregation.

\subsection{Local Spatial Motion Incoherence (LSMI)}

To capture whether temporal changes are spatially coordinated, we define a feature-space motion field and measure its local spatial disagreement. We first reshape tokens into a grid $P_t \in \mathbb{R}^{H_p \times W_p \times D}$ and compute the per-step motion field:
\begin{equation}
    M_t = P_{t+1} - P_t \in \mathbb{R}^{H_p \times W_p \times D}, \quad t=1,\dots,T-1.
\end{equation}
We then compute local spatial gradients of motion using forward differences:
\begin{align}
    \nabla_x M_t(i,j) &= M_t(i,j) - M_t(i,j+1), \quad j=1,\dots,W_p-1,\\
    \nabla_y M_t(i,j) &= M_t(i,j) - M_t(i+1,j), \quad i=1,\dots,H_p-1.
\end{align}
LSMI is the mean magnitude of these gradients, averaged across time and spatial locations:
\begin{equation}
    \mathbf{LSMI} =
    \frac{1}{2}\left(
    \mathbb{E}_{t,i,j}\big[\|\nabla_x M_t(i,j)\|_2\big]
    +
    \mathbb{E}_{t,i,j}\big[\|\nabla_y M_t(i,j)\|_2\big]
    \right).
\end{equation}
Intuitively, real motion tends to be spatially coherent, with neighboring patches changing similarly, whereas localized generation or editing artifacts often yield spatially inconsistent temporal changes, thereby increasing LSMI.

\subsection{Gamma-corrected Fusion and Scoring}

We fuse the two signals with a multiplicative rule:
\begin{equation}
    s(x) = \mathbf{TTR}^{\gamma}\cdot \mathbf{LSMI},
\end{equation}
where larger scores indicate higher likelihood of manipulation. Here $\gamma$ is a fixed, non-learned sharpening constant that amplifies subtle TTR differences. We selected $\gamma=8$ once on a small held-out validation split and then kept it fixed for all benchmarks and FPR targets.

To meet service-aligned constraints, we choose a threshold $\tau$ using real-only threshold calibration such that the resulting FPR on real calibration data matches a target (\eg, $0.1\%$, $1\%$, or $5\%$). At test time, we predict manipulated if $s(x) \ge \tau$ and real otherwise.

\section{Experiments}
\label{sec:experiments}
\subsection{Experimental Setup}
\label{sec:experimental_setup}

\paragraph{Training Data.}
\framework{} is training-free and therefore requires no training dataset. For comparison with learning-based detectors, we train all supervised baselines following the experimental settings of DeMamba~\cite{chen2024demamba} and D3~\cite{zheng2025d3a} using GenVideo-100K~\cite{chen2024demamba} as the sole training set.

\paragraph{Test Benchmarks.}
We evaluate on three benchmarks. FakeParts~\cite{liu2025fakeparts} targets partially edited videos with eight manipulation categories including T2V, TI2V, interpolation, extrapolation, inpainting, outpainting, faceswap, and style change. GenVideo~\cite{chen2024demamba} focuses on T2V content from ten earlier generators. ViF-Bench~\cite{li2025skyra} targets newer 2024--2025 systems spanning fourteen T2V and five I2V generators (\eg, Sora-2, Wan2.2, Kling), providing an out-of-distribution test of generalization. Complete lists of generators for each benchmark are provided in the supplementary material.

\paragraph{Baselines.}
We compare \framework{} against the training-free D3~\cite{zheng2025d3a} method as well as ten supervised detectors: three image-level methods (FID~\cite{zheng2024breakinga}, NPR~\cite{tan2024rethinking}, STIL~\cite{gu2021spatiotemporala}) and seven video-level methods (FTCN~\cite{zheng2021exploring}, MINTIME~\cite{coccomini2024mintime}, TALL~\cite{xu2023tall}, XCLIP~\cite{ni2022expanding}, AIGVDet~\cite{bai2025aigenerated}, DeMamba~\cite{chen2024demamba}, ReStraV~\cite{NEURIPS2025_1d9a4375}). NPR, STIL, FTCN, MINTIME, TALL, XCLIP, and DeMamba are reproduced using the official DeMamba repository, while FID, AIGVDet, and ReStraV are trained and evaluated using their respective official implementations. D3 is evaluated using its official repository.

\paragraph{Implementation Details.}
We adopt a pretrained XCLIP-B/16~\cite{ni2022expanding} vision encoder for patch token extraction. Following D3~\cite{zheng2025d3a}, we extract segments of up to 2 seconds, sample frames at 8 fps with uniform intervals, crop 10\% from the longer side, and resize all frames to $224 \times 224$ pixels. All experiments are conducted on an AMD EPYC 9554 64-Core Processor with an NVIDIA RTX A6000 GPU.

\paragraph{Service-Aligned Evaluation Metric.}
Standard metrics such as AUROC do not reflect deployment settings where falsely rejecting authentic user content must be extremely rare. We therefore report Fake Recall at fixed False Positive Rate (FPR) on real videos $\alpha$ for $\alpha \in \{0.1\%, 1\%, 5\%\}$. 
Let $s(x)$ be the detector score (larger means more likely manipulated), and classify a video as fake if $s(x)\ge \tau$.
We choose a threshold $\tau_\alpha$ using a real-only calibration set $\mathcal{R}$ such that the fraction of real videos flagged as fake is at most $\alpha$:
\begin{equation}
\mathrm{FPR}(\tau) \triangleq \Pr_{x\sim \mathcal{R}}\!\big[s(x)\ge \tau\big], 
\qquad 
\tau_\alpha \triangleq \inf\{\tau:\mathrm{FPR}(\tau)\le \alpha\}.
\end{equation}
We then evaluate detection performance on manipulated videos $\mathcal{F}$ at this operating point:
\begin{equation}
\mathrm{Fake\ Recall}@\alpha \triangleq \Pr_{x\sim \mathcal{F}}\!\big[s(x)\ge \tau_\alpha\big].
\end{equation}

\paragraph{Cross-Real Threshold Transfer.}
To measure robustness to the choice of real calibration source, we calibrate thresholds on two disjoint real sets used in prior work: ROVI~\cite{wu2024languagedriven} (the real subset of FakeParts) and MSR-VTT~\cite{xu2016msrvtt} (the real subset of GenVideo). For each FPR target $\alpha$, we compute $\tau_{\text{ROVI}}$ and $\tau_{\text{MSR-VTT}}$ separately, apply each threshold to the test benchmarks, and report the harmonic mean of the resulting Fake Recall values. This prevents over-crediting methods that are well-calibrated on only one real domain. Per-real-set results and conventional AUROC metrics are provided in the supplementary material.

\begin{table}[p]
\centering

\caption{Fake Recall (\%) on FakeParts under cross-real threshold transfer at FPR $\in \{0.1\%,\, 1\%,\, 5\%\}$. Methods are grouped by horizontal lines. Overall denotes the simple average across all categories. Best results are in \textbf{bold}.}
\label{tab:main_results}
\renewcommand{\arraystretch}{0.95}
\setlength{\tabcolsep}{4pt}

\resizebox{\textwidth}{!}{%
\begin{tabular}{lccccccccccccccc}
\toprule
\multirow{2}{*}{Method}
 & \multicolumn{3}{c}{Extrapolation} & \multicolumn{3}{c}{Faceswap} & \multicolumn{3}{c}{Inpainting} & \multicolumn{3}{c}{Interpolation} & \multicolumn{3}{c}{Outpainting} \\
\cmidrule(lr){2-4} \cmidrule(lr){5-7} \cmidrule(lr){8-10} \cmidrule(lr){11-13} \cmidrule(lr){14-16}
 & 0.1\% & 1\% & 5\% & 0.1\% & 1\% & 5\% & 0.1\% & 1\% & 5\% & 0.1\% & 1\% & 5\% & 0.1\% & 1\% & 5\% \\
\midrule
NPR & 0.00 & 1.03 & 8.43 & 0.00 & 0.00 & 0.53 & 0.12 & 2.75 & 12.45 & 0.00 & 0.02 & 1.28 & 0.00 & 0.00 & 0.16 \\
STIL & 0.09 & 0.18 & 2.79 & 0.08 & 0.16 & 2.40 & 0.69 & 1.36 & 7.56 & 0.17 & 0.48 & 2.58 & 0.18 & 0.34 & 1.96 \\
FID & 0.00 & 0.70 & 5.58 & 0.16 & 8.70 & 35.39 & 0.33 & 6.48 & 28.19 & 1.63 & 10.84 & 34.24 & 6.44 & 28.19 & 52.72 \\
\midrule
MINTIME & 0.04 & 0.33 & 2.63 & 0.00 & 0.03 & 0.78 & 0.06 & 0.82 & 3.53 & 2.17 & 11.67 & 29.55 & 0.07 & 0.50 & 2.60 \\
FTCN & 0.23 & 1.35 & 3.69 & 0.00 & 0.24 & 1.49 & 0.05 & 0.68 & 3.54 & 1.25 & 6.10 & 15.03 & 0.00 & 0.64 & 2.05 \\
TALL & 0.05 & 1.27 & 5.22 & 0.00 & 0.13 & 0.88 & 0.26 & 2.04 & 8.32 & 0.24 & 1.59 & 7.29 & 0.19 & 1.09 & 5.08 \\
XCLIP & 0.00 & 0.03 & 0.70 & 0.00 & 0.12 & 1.36 & 0.08 & 0.78 & 4.92 & 0.18 & 1.03 & 4.72 & 0.09 & 0.45 & 2.18 \\
AIGVDet & 0.42 & 1.77 & 4.87 & 0.89 & 6.47 & 14.63 & 0.98 & 7.10 & 19.79 & 0.43 & 3.18 & 7.62 & 0.56 & 1.80 & 3.50 \\
DeMamba & 0.00 & 0.05 & 0.45 & 0.03 & 0.21 & 1.92 & 0.07 & 1.06 & 5.01 & 0.00 & 1.35 & 8.40 & 0.14 & 1.32 & 6.06 \\
ReStraV & 23.27 & 61.48 & 81.58 & 12.98 & 50.72 & 77.63 & \textbf{19.09} & \textbf{30.70} & \textbf{34.53} & 0.18 & 11.72 & 50.69 & 91.45 & 98.92 & 99.79 \\
\midrule
D3 & 0.00 & 16.77 & 58.66 & 0.04 & 40.86 & 86.89 & 0.00 & 1.20 & 7.05 & 0.00 & 11.60 & 53.61 & 2.72 & 66.03 & 94.13 \\
\framework{} & \textbf{97.86} & \textbf{99.37} & \textbf{99.86} & \textbf{60.09} & \textbf{84.21} & \textbf{96.35} & 2.48 & 14.63 & 33.56 & \textbf{67.81} & \textbf{87.57} & \textbf{96.58} & \textbf{96.52} & \textbf{99.23} & \textbf{99.85} \\
\bottomrule
\end{tabular}%
}

\resizebox{\textwidth}{!}{%
\begin{tabular}{lccccccccccccc}
\toprule
\multirow{2}{*}{Method} & \multicolumn{3}{c}{Style Change} & \multicolumn{3}{c}{T2V} & \multicolumn{3}{c}{TI2V} & & \multicolumn{3}{c}{Overall} \\
\cmidrule(lr){2-4} \cmidrule(lr){5-7} \cmidrule(lr){8-10} \cmidrule(lr){12-14}
 & 0.1\% & 1\% & 5\% & 0.1\% & 1\% & 5\% & 0.1\% & 1\% & 5\% & & 0.1\% & 1\% & 5\% \\
\midrule
NPR & 0.04 & 1.36 & 4.78 & 0.38 & 4.25 & 18.45 & 0.45 & 4.67 & 21.43 & & 0.12 & 1.76 & 8.44 \\
STIL & 2.19 & 4.20 & 15.63 & 6.72 & 11.30 & 31.79 & 3.52 & 6.85 & 22.96 & & 1.70 & 3.11 & 10.96 \\
FID & 12.26 & 38.59 & 64.28 & 11.15 & 37.61 & 62.68 & 5.67 & 30.49 & 63.15 & & 4.71 & 20.20 & 43.28 \\
\midrule
MINTIME & 3.54 & 20.74 & 41.95 & 2.19 & 12.99 & 30.92 & 4.24 & 25.27 & 56.10 & & 1.54 & 9.04 & 21.01 \\
FTCN & 13.98 & 33.91 & 55.65 & 8.57 & 24.13 & 42.81 & 19.41 & 50.10 & 72.75 & & 5.44 & 14.65 & 24.63 \\
TALL & 0.30 & 1.65 & 6.53 & 0.32 & 1.89 & 6.84 & 0.05 & 0.71 & 3.39 & & 0.18 & 1.30 & 5.44 \\
XCLIP & 1.56 & 8.65 & 26.55 & 3.91 & 14.88 & 34.28 & 9.53 & 29.74 & 58.89 & & 1.92 & 6.96 & 16.70 \\
AIGVDet & 3.69 & 8.54 & 12.26 & 31.65 & 47.88 & 56.74 & 34.39 & 61.26 & 76.42 & & 9.13 & 17.25 & 24.48 \\
DeMamba & 1.31 & 8.76 & 25.52 & 1.59 & 9.47 & 24.13 & 3.37 & 17.61 & 43.48 & & 0.81 & 4.98 & 14.37 \\
ReStraV & 18.60 & 69.91 & 90.06 & 46.65 & 79.45 & 90.37 & 73.13 & 92.26 & 97.40 & & 35.67 & 61.89 & 77.75 \\
\midrule
D3 & 0.34 & 40.14 & 79.63 & 14.46 & 68.30 & \textbf{93.22} & 0.66 & 54.49 & 91.56 & & 2.28 & 37.42 & 70.59 \\
\framework{} & \textbf{98.80} & \textbf{99.71} & \textbf{99.92} & \textbf{76.74} & \textbf{85.34} & 91.64 & \textbf{95.27} & \textbf{98.61} & \textbf{99.86} & & \textbf{74.45} & \textbf{83.58} & \textbf{89.70} \\
\bottomrule
\end{tabular}%
}
\end{table}

\begin{table}[p]
\centering
\caption{Fake Recall (\%) on GenVideo under cross-real threshold transfer at FPR $\in \{0.1\%,\, 1\%,\, 5\%\}$.}  
\label{tab:experiment_results}
\renewcommand{\arraystretch}{0.95}
\setlength{\tabcolsep}{3pt}

\resizebox{\textwidth}{!}{%
\begin{tabular}{lcccccccccccccccccc}
\toprule
\multirow{2}{*}{Method}
 & \multicolumn{3}{c}{Crafter~\cite{chen2023videocrafter1opendiffusionmodels}} & \multicolumn{3}{c}{Gen2~\cite{Esser_2023_ICCV}} & \multicolumn{3}{c}{HotShot~\cite{Mullan_Hotshot-XL_2023}} & \multicolumn{3}{c}{Lavie~\cite{wang2025lavie}} & \multicolumn{3}{c}{ModelScope~\cite{wang2023modelscopetexttovideotechnicalreport}} & \multicolumn{3}{c}{MoonValley~\cite{moonvalley2022moonvalley}} \\
\cmidrule(lr){2-4} \cmidrule(lr){5-7} \cmidrule(lr){8-10} \cmidrule(lr){11-13} \cmidrule(lr){14-16} \cmidrule(lr){17-19}
 & 0.1\% & 1\% & 5\% & 0.1\% & 1\% & 5\% & 0.1\% & 1\% & 5\% & 0.1\% & 1\% & 5\% & 0.1\% & 1\% & 5\% & 0.1\% & 1\% & 5\% \\
\midrule
NPR & 0.42 & 7.61 & 30.60 & 0.28 & 3.27 & 15.14 & 0.00 & 1.68 & 20.25 & 0.00 & 0.84 & 9.56 & 1.64 & 8.88 & 23.97 & 0.63 & 4.53 & 25.30 \\
STIL & 22.75 & 31.19 & 60.45 & 12.83 & 20.49 & 53.42 & 12.71 & 21.59 & 46.23 & 10.50 & 16.46 & 38.78 & 10.14 & 17.10 & 43.23 & 20.61 & 31.22 & 64.96 \\
FID & 78.71 & 93.76 & 98.23 & 17.47 & 54.61 & 80.12 & 56.46 & 90.16 & 97.91 & 29.09 & 69.23 & 89.30 & 22.88 & 64.27 & 87.77 & 63.23 & 91.98 & 98.22 \\
\midrule
MINTIME & 15.06 & 59.57 & 90.95 & 9.10 & 51.27 & 85.51 & 1.49 & 10.46 & 30.85 & 13.23 & 56.53 & 85.93 & 6.81 & 29.50 & 58.57 & 9.42 & 43.20 & 79.71 \\
FTCN & 61.09 & 88.03 & 96.60 & 43.44 & 77.09 & 90.16 & 15.56 & 42.77 & 66.90 & 33.32 & 68.78 & 86.64 & 21.76 & 48.92 & 69.52 & 30.80 & 71.41 & 91.26 \\
TALL & 0.46 & 2.26 & 6.70 & 0.22 & 2.49 & 9.00 & 0.84 & 3.27 & 10.13 & 0.00 & 0.77 & 3.54 & 0.41 & 3.30 & 8.94 & 0.00 & 1.00 & 4.88 \\
XCLIP & 12.00 & 42.72 & 74.03 & 5.98 & 28.32 & 68.55 & 5.53 & 18.99 & 48.72 & 5.78 & 18.73 & 45.42 & 5.98 & 23.47 & 58.17 & 7.78 & 30.57 & 70.79 \\
AIGVDet & 79.71 & 93.15 & 97.26 & 64.78 & 81.91 & 90.58 & 0.00 & 0.00 & 0.00 & 15.39 & 31.14 & 44.37 & 53.99 & 77.86 & 89.40 & \textbf{87.62} & \textbf{96.50} & \textbf{98.47} \\
DeMamba & 21.94 & 54.77 & 79.07 & 22.96 & 59.21 & 83.98 & 4.08 & 21.31 & 45.52 & 3.74 & 22.57 & 50.95 & 3.89 & 20.40 & 46.27 & 2.86 & 16.72 & 40.36 \\
ReStraV & 20.12 & 67.68 & 90.78 & \textbf{64.86} & 84.82 & 93.07 & 42.47 & 93.37 & 98.84 & 31.00 & 74.97 & 88.58 & 59.37 & 86.48 & 94.19 & 55.44 & 80.02 & 87.01 \\
\midrule
D3 & 15.05 & 75.72 & 94.00 & 8.39 & \textbf{87.91} & \textbf{98.73} & 3.05 & 61.21 & 89.38 & 2.24 & 65.21 & 88.34 & 1.96 & 53.09 & 84.63 & 19.41 & 85.35 & 97.18 \\
\framework{} & \textbf{91.65} & \textbf{97.49} & \textbf{99.71} & 60.49 & 81.73 & 94.96 & \textbf{99.93} & \textbf{100.00} & \textbf{100.00} & \textbf{93.80} & \textbf{96.96} & \textbf{98.82} & \textbf{89.01} & \textbf{94.56} & \textbf{98.14} & 57.22 & 81.43 & 95.77 \\
\bottomrule
\end{tabular}%
}

\resizebox{\textwidth}{!}{%
\begin{tabular}{lcccccccccccccccc}
\toprule
\multirow{2}{*}{Method} & \multicolumn{3}{c}{MorphStudio~\cite{morph2023morphstudio}} & \multicolumn{3}{c}{Show\_1~\cite{zhang2025show1}} & \multicolumn{3}{c}{Sora~\cite{brooks2024video}} & \multicolumn{3}{c}{WildScrape~\cite{chen2024demamba}} & & \multicolumn{3}{c}{Overall} \\
\cmidrule(lr){2-4} \cmidrule(lr){5-7} \cmidrule(lr){8-10} \cmidrule(lr){11-13} \cmidrule(lr){15-17}
 & 0.1\% & 1\% & 5\% & 0.1\% & 1\% & 5\% & 0.1\% & 1\% & 5\% & 0.1\% & 1\% & 5\% & & 0.1\% & 1\% & 5\% \\
\midrule
NPR & 0.28 & 2.74 & 17.94 & 0.00 & 0.00 & 0.82 & 0.00 & 0.00 & 16.48 & 0.22 & 3.25 & 15.57 & & 0.35 & 3.28 & 17.56 \\
STIL & 15.29 & 21.80 & 49.91 & 8.14 & 14.37 & 40.08 & 7.14 & 12.24 & 33.01 & 7.22 & 11.56 & 30.99 & & 12.73 & 19.80 & 46.11 \\
FID & 29.10 & 71.67 & 91.75 & 18.19 & 56.37 & 86.94 & 0.00 & 8.83 & 25.25 & 25.17 & 46.08 & 66.00 & & 34.03 & 64.69 & 82.15 \\
\midrule
MINTIME & 14.03 & 55.51 & 88.93 & 6.99 & 36.54 & 73.00 & 2.38 & 14.06 & 28.57 & 13.00 & 43.49 & 67.29 & & 9.15 & 40.01 & 68.93 \\
FTCN & 55.20 & 82.67 & 94.62 & 20.81 & 53.68 & 75.76 & 10.42 & 18.99 & 33.01 & 35.55 & 55.63 & 67.50 & & 32.80 & 60.80 & 77.20 \\
TALL & 0.78 & 2.95 & 9.39 & 0.00 & 2.82 & 9.52 & 0.00 & 0.00 & 2.68 & 0.17 & 1.03 & 5.90 & & 0.29 & 1.99 & 7.07 \\
XCLIP & 19.04 & 43.27 & 71.59 & 6.73 & 28.76 & 54.24 & 5.10 & 17.86 & 43.30 & 8.34 & 24.11 & 48.90 & & 8.23 & 27.68 & 58.37 \\
AIGVDet & 68.36 & 86.08 & 94.58 & 11.58 & 24.31 & 34.16 & 37.72 & 67.81 & 79.37 & 34.64 & 46.88 & 55.78 & & 45.38 & 60.56 & 68.40 \\
DeMamba & 18.61 & 47.15 & 73.28 & 10.58 & 38.06 & 67.43 & 2.86 & 12.61 & 27.42 & 8.50 & 26.35 & 45.90 & & 10.00 & 31.91 & 56.02 \\
ReStraV & 38.15 & 81.76 & 93.72 & 40.09 & 87.82 & 96.83 & 64.67 & 89.11 & 96.30 & 39.07 & 72.82 & 83.88 & & 45.52 & 81.89 & 92.32 \\
\midrule
D3 & 2.78 & 59.61 & 89.99 & 1.40 & 72.99 & 95.73 & 0.00 & 64.56 & 94.61 & 0.44 & 39.31 & 72.63 & & 5.47 & 66.50 & 90.52 \\
\framework{} & \textbf{96.19} & \textbf{98.86} & \textbf{99.79} & \textbf{98.13} & \textbf{99.93} & \textbf{100.00} & \textbf{90.93} & \textbf{99.10} & \textbf{100.00} & \textbf{74.30} & \textbf{80.89} & \textbf{86.27} & & \textbf{85.16} & \textbf{93.09} & \textbf{97.35} \\
\bottomrule
\end{tabular}%
}
\end{table}

\begin{table}[t]
\centering
\caption{Fake Recall (\%) on ViF-Bench under cross-real threshold transfer at FPR $= 0.1\%$. Results for FPR $\in \{1\%,\, 5\%\}$ are provided in the supplementary material.}
\label{tab:vif_results}
\renewcommand{\arraystretch}{1.0}

\resizebox{\textwidth}{!}{%
\begin{tabular}{lcccccccccccccc|ccccc|c}
\toprule
\multirow{2}{*}{Method} & \multicolumn{14}{c}{T2V Models} & \multicolumn{5}{c}{I2V Models} & \multirow{2}{*}{Overall} \\
\cmidrule(lr){2-15} \cmidrule(lr){16-20}
 & \rotatebox{90}{CogX1.5~\cite{yang2025cogvideox}} & \rotatebox{90}{Hunyuan~\cite{kong2025hunyuanvideosystematicframeworklarge}} & \rotatebox{90}{LTX-T~\cite{hacohen2024ltxvideorealtimevideolatent}} & \rotatebox{90}{SkyV2~\cite{chen2025skyreelsv2infinitelengthfilmgenerative}} & \rotatebox{90}{Wan2.1~\cite{wan2025wana}} & \rotatebox{90}{Wan2.1V~\cite{wan2025wana}} & \rotatebox{90}{Wan2.2~\cite{wan2025wana}} & \rotatebox{90}{Wan2.2TI-T~\cite{wan2025wana}} & \rotatebox{90}{Gen4~\cite{runway}} & \rotatebox{90}{Hailuo-02~\cite{hailuo}} & \rotatebox{90}{Kling-V1~\cite{klinga}} & \rotatebox{90}{Pika-V2~\cite{pika}} & \rotatebox{90}{Pix4.5~\cite{pixverse}} & \rotatebox{90}{Sora-2~\cite{openai2025sora}} & \rotatebox{90}{Hunyuan-I~\cite{kong2025hunyuanvideosystematicframeworklarge}} & \rotatebox{90}{LTX-I~\cite{hacohen2024ltxvideorealtimevideolatent}} & \rotatebox{90}{SkyV2-I~\cite{chen2025skyreelsv2infinitelengthfilmgenerative}} & \rotatebox{90}{Wan2.2-I~\cite{wan2025wana}} & \rotatebox{90}{Wan2.2TI-I~\cite{wan2025wana}} &  \\
\midrule
NPR & 0.00 & 0.00 & 0.00 & 0.00 & 0.00 & 0.00 & 0.00 & 0.00 & 0.00 & 0.00 & 0.00 & 0.00 & 0.00 & 0.00 & 0.00 & 0.00 & 0.00 & 0.00 & 0.00 & 0.00 \\
STIL & 2.02 & 3.05 & 5.58 & 3.54 & 3.03 & 0.51 & 6.57 & 1.52 & 1.32 & 3.51 & 1.71 & 1.08 & 2.21 & 1.11 & 1.01 & 1.52 & 0.51 & 1.01 & 0.00 & 2.15 \\
FID & 10.59 & 5.89 & 0.96 & 6.09 & 9.84 & 0.97 & 3.91 & 0.91 & 10.21 & 5.88 & 1.07 & 21.54 & 4.63 & 0.93 & 0.00 & 5.10 & 2.39 & 7.41 & 0.93 & 5.22 \\
\midrule
MINTIME & 11.10 & 10.71 & 7.44 & 10.66 & 15.38 & 2.95 & 10.88 & 2.96 & 9.25 & 19.58 & 9.39 & 17.86 & 14.29 & 2.71 & 0.00 & 3.65 & 2.13 & 0.81 & 0.85 & 8.03 \\
FTCN & 18.61 & 24.57 & 14.07 & 23.43 & 16.39 & 6.17 & 16.72 & 1.68 & 16.81 & 42.84 & 22.26 & 28.46 & 39.23 & 3.45 & 0.00 & 7.84 & 4.02 & 4.71 & 2.50 & 15.46 \\
TALL & 0.00 & 0.51 & 0.00 & 0.00 & 0.00 & 0.00 & 0.00 & 0.00 & 0.00 & 0.00 & 0.00 & 0.00 & 0.00 & 0.00 & 0.00 & 0.00 & 0.00 & 0.00 & 0.51 & 0.05 \\
XCLIP & 1.21 & 2.71 & 0.00 & 3.37 & 0.00 & 0.51 & 2.42 & 0.00 & 0.00 & 2.63 & 0.57 & 0.00 & 0.55 & 0.00 & 0.00 & 1.01 & 0.00 & 0.00 & 0.00 & 0.79 \\
AIGVDet & 21.04 & 27.76 & 46.82 & 57.24 & 28.51 & 24.69 & 42.14 & 16.32 & 11.47 & 50.35 & 18.29 & 31.32 & 47.51 & 18.66 & 0.92 & 13.30 & 10.39 & 15.20 & 6.13 & 25.69 \\
DeMamba & 1.73 & 7.41 & 0.76 & 3.52 & 5.37 & 1.44 & 5.31 & 2.69 & 4.85 & 9.93 & 2.85 & 1.62 & 5.87 & 1.67 & 0.68 & 1.62 & 1.53 & 2.34 & 1.44 & 3.30 \\
ReStraV & 17.35 & 36.31 & 23.22 & 42.81 & 36.97 & 28.47 & 35.59 & 20.63 & 25.08 & 48.00 & 51.70 & 45.72 & 34.63 & 19.95 & 32.32 & 34.25 & 48.16 & 46.94 & 16.58 & 33.93 \\
\midrule
D3 & 14.32 & 0.00 & 2.01 & 0.00 & 0.00 & 0.00 & 0.00 & 0.00 & 0.00 & 0.00 & 1.13 & 0.00 & 0.00 & 0.00 & 1.01 & 0.00 & 0.00 & 0.00 & 0.00 & 0.97 \\
\framework{} & \textbf{84.77} & \textbf{80.60} & \textbf{73.98} & \textbf{73.79} & \textbf{92.05} & \textbf{93.37} & \textbf{80.71} & \textbf{96.66} & \textbf{87.38} & \textbf{70.53} & \textbf{93.23} & \textbf{97.01} & \textbf{71.42} & \textbf{91.70} & \textbf{61.58} & \textbf{82.38} & \textbf{83.09} & \textbf{92.64} & \textbf{98.98} & \textbf{84.52} \\
\bottomrule
\end{tabular}%
}
\end{table}

\subsection{Results and Analysis}
\label{sec:result_analysis}

Fake Recall under cross-real threshold transfer is reported in \cref{tab:main_results,tab:experiment_results,tab:vif_results}. Across all three benchmarks, \framework{} consistently achieves the highest Fake Recall despite being entirely training-free.

\paragraph{FakeParts.}
\framework{} attains 74.45\% Fake Recall at FPR\,=\,0.1\% overall (95\% category-stratified bootstrap CI $[72.42\%,\,76.14\%]$; see supplementary material), compared to 35.67\% for the best supervised baseline (ReStraV) and 2.28\% for D3 (CI $[0.53\%,\,5.43\%]$). Several categories are near-saturated even at this strictest threshold: Extrapolation (97.86\%), Outpainting (96.52\%), Style Change (98.80\%), and TI2V (95.27\%). Faceswap (60.09\%) and Interpolation (67.81\%) also show substantial improvements over all baselines. Inpainting remains the most difficult category (2.48\% at FPR\,=\,0.1\%). Nevertheless, \framework{} ranks second among the compared methods, reflecting the challenge of detecting edits confined to a small spatial region per frame.

\paragraph{GenVideo.}
\framework{} achieves 85.16\% Fake Recall at FPR\,=\,0.1\%, rising to 93.09\% at 1\% and 97.35\% at 5\%. The strongest supervised competitor under the strictest constraint is ReStraV at 45.52\%, while D3 reaches only 5.47\%. \framework{} exceeds 90\% at FPR\,=\,0.1\% on six of the ten generators, with performance improves substantially as FPR relaxes, consistent with a score distribution that provides stable separation between real and generated content.

\paragraph{ViF-Bench.}
At FPR\,=\,0.1\%, \framework{} achieves 84.52\% overall Fake Recall and ranks first on every individual generator. Supervised detectors degrade sharply: the best baseline (ReStraV) reaches 33.93\%, while D3 drops to 0.97\%. This confirms that patch-level TTR and LSMI cues capture generation artifacts that persist across architectural generations of video synthesizers, without requiring retraining.

\paragraph{Analysis.}
The performance gap is most pronounced at FPR\,=\,0.1\%, precisely the regime relevant to deployment. This supports our central hypothesis: localized spatiotemporal incoherence---measured at the patch level on frozen encoder tokens---provides a discriminative and broadly generalizable signal. By avoiding clip-level aggregation, \framework{} preserves sensitivity to partial manipulations while remaining robust across diverse and newly emerging video generators, without any generator-specific training or adaptation, under real-only threshold calibration with cross-real transfer.

\begin{figure}[t]
  \centering
  \begin{minipage}[t]{0.48\linewidth}
    \centering
    \captionof{table}{Cross-real threshold transfer stability. Actual FPR (\%) on a held-out real domain when the threshold is calibrated to a target FPR on a disjoint real domain. Values closer to the nominal target (0.1, 1, or 5\%) indicate more stable cross-domain calibration.}
    \label{tab:cross_real_fpr}
    \resizebox{\linewidth}{!}{%
    \begin{tabular}{l ccc ccc}
      \toprule
      & \multicolumn{3}{c}{MSR-VTT $\rightarrow$ ROVI} & \multicolumn{3}{c}{ROVI $\rightarrow$ MSR-VTT} \\
      \cmidrule(lr){2-4} \cmidrule(lr){5-7}
      Method & @0.1 & @1 & @5 & @0.1 & @1 & @5 \\
      \midrule
      FID      & 0.62 & 8.25 & 38.64 & 0.01 & 0.16 & 0.67 \\
      FTCN     & 0.48 & 2.28 & 7.40 & 0.01 & 0.28 & 2.68 \\
      AIGVDet  & 13.22 & 25.12 & 31.69 & 0.00 & 0.00 & 0.02 \\
      ReStraV  & 0.00 & 0.00 & 0.00 & 15.30 & 39.86 & 66.10 \\
      D3       & 0.00 & 0.07 & 1.52 & 1.18 & 4.06 & 9.04 \\
      \midrule
      \framework{}     & 0.03 & 0.29 & 4.22 & 0.56 & 2.18 & 5.49 \\
      \bottomrule
    \end{tabular}%
    }
  \end{minipage}
  \hfill
  \begin{minipage}[t]{0.50\linewidth}
    \vspace{0pt}
    \centering
    \includegraphics[width=\linewidth]{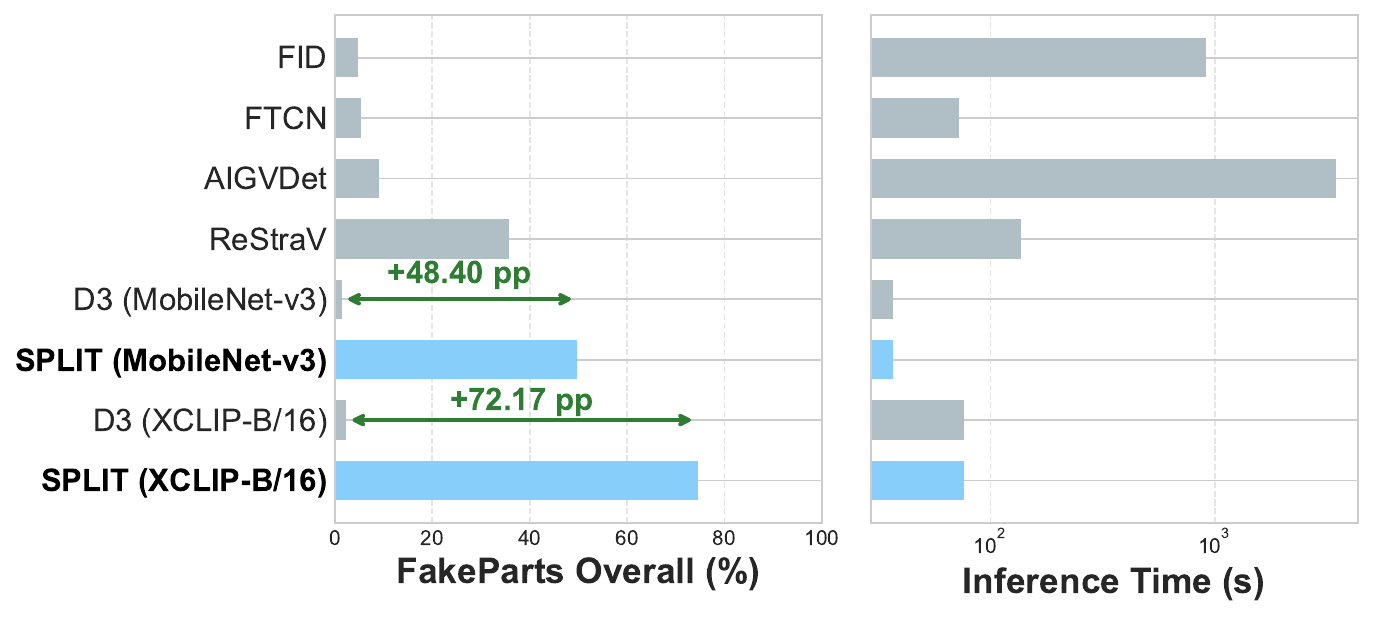}
    \captionof{figure}{Comparison of FakeParts Overall Fake Recall at FPR = 0.1\% (left) and end-to-end inference time on 1,000 FakeParts T2V samples and batch size of 1 (right, log scale). Inference time for AIGVDet includes RAFT~\cite{teed2020raft}-based optical flow preprocessing.}
    \label{fig:speed}
  \end{minipage}
\end{figure}
\subsection{Cross-Real Threshold Transfer}
\label{sec:cross_real_transfer}

A practical deployment concern is whether a threshold calibrated on one real-only source remains reliable when the target real distribution differs. As shown in \cref{tab:cross_real_fpr}, we report the actual rejection rate on a held-out real domain when the threshold is set to achieve a target FPR on a different real domain. We compare against the five strongest competing methods selected by FakeParts Overall Fake Recall at FPR\,=\,0.1\%.

Our method maintains the transferred FPR close to the intended target in both directions, indicating stable calibration across real domains rather than over-specialization to a particular real set. Several baselines, by contrast, exhibit pronounced domain-dependent drift: a threshold tuned on one real source can become substantially more permissive or more restrictive on the other, undermining  operation at ultra-low FPR. This stability, combined with the strong Fake Recall reported in \S\ref{sec:result_analysis}, confirms that \framework{}'s patch-level signals generalize not only across generators but also across real-data distributions---a prerequisite for practical threshold setting without access to target-domain real samples.

\subsection{Inference Efficiency}
\label{sec:efficiency}

Inference time on 1{,}000 FakeParts T2V samples is reported in \cref{fig:speed}. Because feature encoding dominates the computational cost, our lightweight patch-level TTR and LSMI computations add negligible overhead to the frozen encoder forward pass. When paired with the same backbone, \framework{} matches D3's runtime while boosting the FakeParts Overall Fake Recall at FPR\,=\,0.1\% from 2.28\% to 74.45\% with XCLIP-B/16. \framework{} delivers substantially stronger detection under service-aligned thresholds without sacrificing practical throughput.

\begin{table}[tb]
\centering
\caption{Fake Recall (\%) on FakeParts Overall at FPR\,$\in\{0.1\%,\,1\%,\,5\%\}$ for Gaussian blur ($\sigma\!=\!1,2$), JPEG compression ($q\!=\!90,80$), and Axis Flips ($x,y$) post-processing, compared against the unperturbed baseline.}
\label{tab:robustness_results}
\renewcommand{\arraystretch}{1.0}
\setlength{\tabcolsep}{1.5pt}

\resizebox{\textwidth}{!}{%
\begin{tabular}{lccccccccccccccccccccc}
\toprule
\multirow{2}{*}{Method} 
 & \multicolumn{3}{c}{Baseline} & \multicolumn{3}{c}{Gaussian Blur ($\sigma=1$)} & \multicolumn{3}{c}{Gaussian Blur ($\sigma=2$)} & \multicolumn{3}{c}{JPEG ($q=90$)} & \multicolumn{3}{c}{JPEG ($q=80$)} & \multicolumn{3}{c}{Flip ($x$-axis) } & \multicolumn{3}{c}{Flip ($y$-axis)} \\
\cmidrule(lr){2-4} \cmidrule(lr){5-7} \cmidrule(lr){8-10} \cmidrule(lr){11-13} \cmidrule(lr){14-16} \cmidrule(lr){17-19} \cmidrule(lr){20-22}
 & 0.1\% & 1\% & 5\% & 0.1\% & 1\% & 5\% & 0.1\% & 1\% & 5\% & 0.1\% & 1\% & 5\% & 0.1\% & 1\% & 5\% & 0.1\% & 1\% & 5\% & 0.1\% & 1\% & 5\% \\
\midrule
NPR      & 0.12 & 1.76 & 8.44  & 0.25 & 2.10 & 9.49  & 0.24 & 1.96 & 6.67  & 0.11 & 1.66 & 8.24  & 0.13 & 1.70 & 8.29  & 0.12 & 1.77 & 9.03 & 0.14 & 1.79 & 8.78 \\
STIL     & 1.70 & 3.11 & 10.96 & 1.41 & 3.35 & 11.58 & 1.08 & 2.67 & 9.70  & 1.68 & 2.97 & 10.52 & 1.69 & 3.15 & 10.24 & 1.21 & 3.36 & 11.33 & 1.67 & 3.05 & 10.01 \\
FID      & 4.71 & 20.20 & 43.28 & 3.83 & 11.63 & 24.72 & 1.48 & 6.50 & 18.68 & 4.78 & 20.22 & 43.68 & 4.98 & 21.82 & 44.11 & 4.20 & 21.25 & 44.78 & 0.22 & 1.18 & 5.58 \\
\midrule
MINTIME & 1.54 & 9.04 & 21.01 & 1.18 & 6.36 & 16.60 & 0.43 & 3.84 & 11.75 & 1.22 & 7.31 & 17.72 & 0.94 & 5.98 & 15.80 & 1.95 & 10.02 & 22.61 & 1.20 & 6.64 & 16.56 \\
FTCN     & 5.44 & 14.65 & 24.63 & 4.74 & 10.68 & 19.97 & 1.50 & 6.21 & 14.19 & 5.48 & 13.70 & 23.33 & 5.74 & 12.62 & 22.10 & 6.13 & 15.95 & 26.53 & 5.18 & 12.81 & 22.93 \\
TALL     & 0.18 & 1.30 & 5.44  & 0.18 & 1.15 & 5.28  & 0.15 & 1.16 & 5.00  & 0.18 & 1.28 & 5.45  & 0.18 & 1.29 & 5.52  & 0.19 & 1.25 & 5.34 & 0.19 & 1.29 & 5.63 \\
XCLIP    & 1.92 & 6.96 & 16.70 & 2.12 & 8.05 & 17.49 & 1.67 & 6.15 & 14.29 & 1.31 & 5.92 & 14.78 & 1.19 & 5.41 & 13.98 & 3.06 & 8.11 & 17.00 & 1.49 & 6.11 & 15.06 \\
AIGVDet  & 9.13 & 17.25 & 24.48 & 9.35 & 18.21 & 25.10 & 9.96 & 18.65 & 26.96 & 9.99 & 16.43 & 24.12 & 9.49 & 16.76 & 24.01 & 9.67 & 17.15 & 24.03 & 8.27 & 16.86 & 23.54 \\
DeMamba  & 0.81 & 4.98 & 14.37 & 1.29 & 4.80 & 13.05 & 0.82 & 3.21 & 9.86  & 1.07 & 4.56 & 13.55 & 1.03 & 4.81 & 13.28 & 1.65 & 6.09 & 15.83 & 0.99 & 5.04 & 15.05 \\
ReStraV  & 35.67 & 61.89 & 77.75 & 44.46 & 72.60 & 82.99 & 54.73 & \textbf{75.13} & 83.33 & 32.66 & 55.99 & 75.08 & 24.42 & 47.32 & 71.83 & 30.25 & 54.85 & 75.48 & 38.50 & 60.50 & 77.55 \\
\midrule
D3       & 2.28 & 37.42 & 70.59 & 1.53 & 28.60 & 66.87 & 1.89 & 23.57 & 63.79 & 2.16 & 38.27 & 71.03 & 2.78 & 39.84 & 71.73 & 2.50 & 40.98 & 74.70 & 2.09 & 37.65 & 70.32 \\
\framework{}     & \textbf{74.45} & \textbf{83.58} & \textbf{89.70} & \textbf{62.54} & \textbf{77.49} & \textbf{88.71} & \textbf{55.42} & 73.50 & \textbf{88.47} & \textbf{70.45} & \textbf{80.89} & \textbf{88.83} & \textbf{64.40} & \textbf{77.11} & \textbf{87.12} & \textbf{75.73} & \textbf{82.79} & \textbf{88.84} & \textbf{66.92} & \textbf{80.42} & \textbf{88.52} \\
\bottomrule
\end{tabular}%
}
\end{table}

\subsection{Robustness to Post-Processing}
\label{sec:robustness}

We evaluate robustness on FakeParts Overall under common post-processing operations---Gaussian blur ($\sigma\!=\!1,2$), JPEG compression ($q\!=\!90,80$), and geometric axis flips (horizontal~$x$, vertical~$y$)---and report Fake Recall at service-aligned FPR targets in \cref{tab:robustness_results}.

Across perturbations and operating points, \framework{} remains the best method by a wide margin. Blur is the most disruptive: at FPR\,=\,0.1\%, Fake Recall drops from 74.45\% to 55.42\% under $\sigma\!=\!2$, yet still exceeds the best supervised baseline (ReStraV, 54.73\%). JPEG compression is substantially less harmful (\framework{} retains 70.45\% at $q\!=\!90$ and 64.40\% at $q\!=\!80$), suggesting our patch-level spatiotemporal cues do not depend on fragile high-frequency pixel artifacts. Geometric flips have minimal impact: horizontal flipping leaves performance essentially unchanged, while vertical flipping causes only a modest decrease (66.92\% at FPR\,=\,0.1\%), consistent with the fact that our scoring aggregates patch-wise evidence without relying on absolute spatial orientation.

At the looser FPR\,=\,5\% operating point, performance remains near-saturated and stable across all perturbations (87.12--89.70\%), indicating that \framework{} maintains robust detection under realistic video transformations while preserving a substantial advantage in the deployment-critical ultra-low-FPR regime.
\section{Ablation Studies}
\label{sec:ablation}

\begin{figure}[t]
  \centering
  \begin{minipage}[t]{0.48\linewidth}
    \centering
    \captionof{table}{Component ablation on FakeParts Overall. Fake Recall (\%) is reported under cross-real threshold transfer at FPR $\in \{0.1\%, 1\%, 5\%\}$. Checkmarks denote active components.}
    \label{tab:ablation_component}
    \resizebox{\linewidth}{!}{%
    \begin{tabular}{cccccc}
    \toprule
    \multirow{2}{*}{Patch-level} & \multirow{2}{*}{TTR} & \multirow{2}{*}{LSMI} & \multicolumn{3}{c}{FakeParts} \\
    \cmidrule(lr){4-6}
     & & & 0.1\% & 1\% & 5\% \\
    \midrule
                 & \checkmark &            & 54.21 & 76.72 & 88.77 \\
    \checkmark   & \checkmark &            & 68.29 & 82.09 & 89.37 \\
    \checkmark   &            & \checkmark & 8.04 & 14.63 & 25.07 \\
    \checkmark   & \checkmark & \checkmark & \textbf{74.45} & \textbf{83.58} & \textbf{89.70} \\
    \bottomrule
    \end{tabular}%
    }
  \end{minipage}
  \hfill
  \begin{minipage}[t]{0.50\linewidth}
    \vspace{0pt} 
    \centering
    \includegraphics[width=\linewidth]{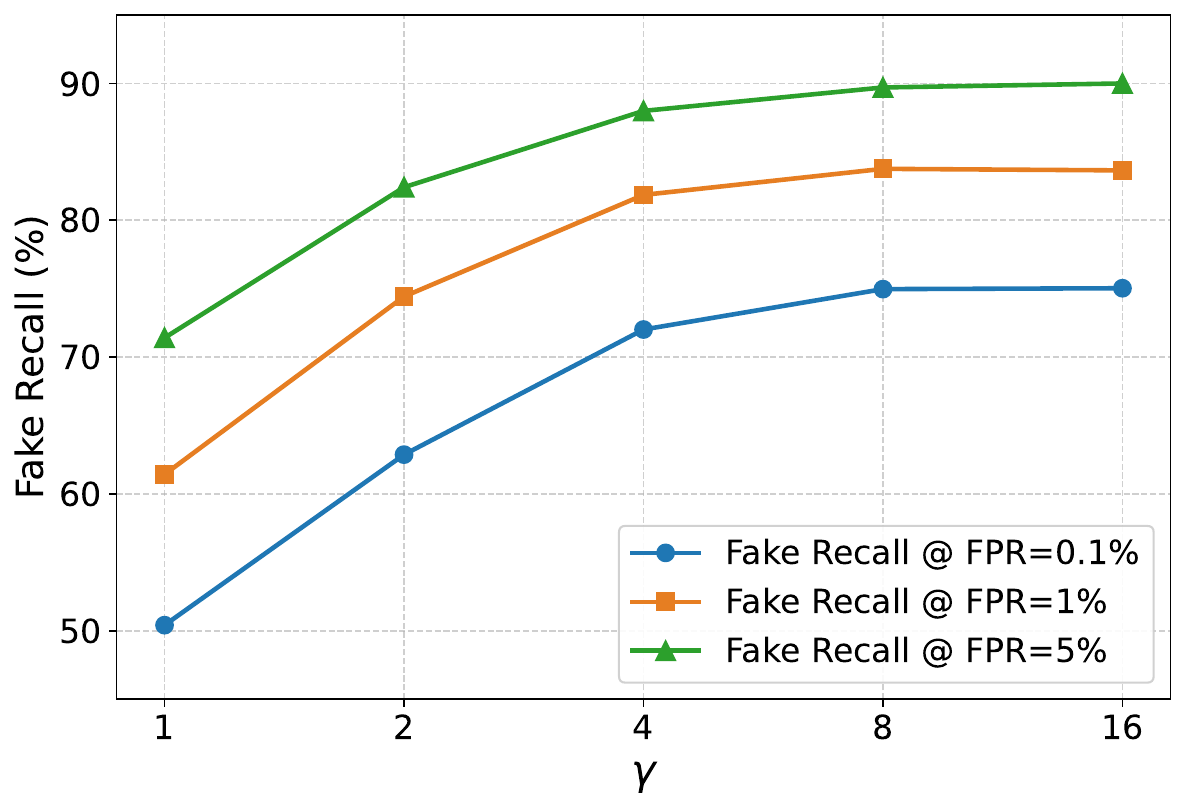}
    \captionof{figure}{Effect of the gamma exponent $\gamma$ on FakeParts Overall Fake Recall (\%).}
    \label{fig:gamma}
  \end{minipage}
\end{figure}

\subsection{Contribution of Each Component}
\label{sec:ablation_component}

We isolate the contribution of each \framework{} component on FakeParts Overall under cross-real threshold transfer, as summarized in \cref{tab:ablation_component}. Replacing global \texttt{[CLS]} embeddings with patch-level tokens for TTR yields the single largest improvement, boosting Fake Recall by +14.08 points at FPR\,=\,0.1\% (from 54.21\% to 68.29\%), confirming that patch-level granularity is essential for detecting partial manipulations. LSMI alone is a weaker standalone signal, yet it captures complementary spatial-coherence evidence: fusing Patch-level TTR with LSMI achieves the best performance across all operating points, with the largest margin at the strictest threshold (+6.16 over Patch-level TTR alone). These results confirm that patch-level temporal roughness is the primary discriminative cue, while LSMI provides additional spatial incoherence information that further strengthens separation under strict real-only calibration. We include an extended analysis of TTR step size and qualitative diagnostics in the supplementary material.

\subsection{Effect of Gamma Exponent $\gamma$}
\label{sec:ablation_gamma}

As shown in \cref{fig:gamma}, we examine the influence of the gamma exponent $\gamma$ in the fusion rule $s(x)=\mathbf{TTR}^{\gamma}\cdot\mathbf{LSMI}$ and report FakeParts Overall Fake Recall under cross-real threshold transfer at FPR $\in\{0.1\%,1\%,5\%\}$. Increasing $\gamma$ from 1 to 8 yields consistent and substantial gains across all operating points, confirming that gamma correction effectively amplifies subtle TTR differences to sharpen the separation between real and manipulated videos. Beyond $\gamma=8$, performance saturates: $\gamma=16$ offers negligible further improvement and can even produce marginal regression at certain FPR levels. We therefore adopt $\gamma=8$ throughout. Additional sensitivity analyses on GenVideo and ViF-Bench show the same trend and are provided in the supplementary material.

\begin{table}[t]
\centering
\caption{Effect of the vision encoder on Fake Recall (\%).}
\label{tab:comparison}
\resizebox{\textwidth}{!}{
\setlength{\tabcolsep}{3.5pt}
\begin{tabular}{@{} ll @{\hspace{8mm}} ccc @{\hspace{6mm}} ccc @{\hspace{6mm}} ccc @{}}
\toprule
\multirow{2}{*}{Vision Encoder} & \multirow{2}{*}{Method} & \multicolumn{3}{c}{FakeParts} & \multicolumn{3}{c}{GenVideo} & \multicolumn{3}{c}{ViF-Bench} \\
\cmidrule(lr){3-5} \cmidrule(lr){6-8} \cmidrule(lr){9-11}
& & 0.1\% & 1\% & 5\% & 0.1\% & 1\% & 5\% & 0.1\% & 1\% & 5\% \\
\midrule
\multirow{2}{*}{CLIP-B/16}
& D3   & 1.66  & 39.79 & 69.94 & 2.37  & 58.19 & 86.13 & 0.74  & 48.62 & 88.78 \\
& \framework{} & \textbf{71.47} & \textbf{81.56} & \textbf{89.39} & \textbf{85.87} & \textbf{93.14} & \textbf{97.81} & \textbf{85.29} & \textbf{94.86} & \textbf{98.38} \\
\midrule
\multirow{2}{*}{CLIP-B/32}
& D3   & 0.64  & 31.17 & 67.19 & 0.14  & 47.89 & 85.39 & 0.50  & 45.53 & 89.50 \\
& \framework{} & \textbf{47.37} & \textbf{72.01} & \textbf{88.56} & \textbf{63.55} & \textbf{82.33} & \textbf{95.54} & \textbf{54.76} & \textbf{87.22} & \textbf{98.12} \\
\midrule
\multirow{2}{*}{DINOv2-B}
& D3   & 4.00  & 36.41 & 61.49 & 16.32 & 64.01 & 84.08 & 3.97  & 54.90 & 83.40 \\
& \framework{} & \textbf{47.77} & \textbf{70.73} & \textbf{87.63} & \textbf{68.03} & \textbf{84.18} & \textbf{95.50} & \textbf{56.64} & \textbf{86.52} & \textbf{97.98} \\
\midrule
\multirow{2}{*}{DINOv2-L}
& D3   & 4.12  & 28.24 & 54.54 & 15.99 & 55.26 & 78.43 & 6.21  & 44.00 & 75.86 \\
& \framework{} & \textbf{49.06} & \textbf{68.25} & \textbf{86.45} & \textbf{68.71} & \textbf{83.32} & \textbf{95.09} & \textbf{55.95} & \textbf{82.64} & \textbf{97.47} \\
\midrule
\multirow{2}{*}{XCLIP-B/16}
& D3   & 2.28  & 37.42 & 70.59 & 5.47  & 66.50 & 90.52 & 0.97  & 46.77 & 87.77 \\
& \framework{} & \textbf{74.45} & \textbf{83.58} & \textbf{89.70} & \textbf{85.16} & \textbf{93.09} & \textbf{97.35} & \textbf{84.52} & \textbf{95.08} & \textbf{98.46} \\
\midrule
\multirow{2}{*}{XCLIP-B/32}
& D3   & 0.67  & 27.63 & 65.75 & 0.94  & 54.34 & 87.43 & 0.53  & 34.43 & 81.72 \\
& \framework{} & \textbf{60.23} & \textbf{77.34} & \textbf{89.07} & \textbf{70.19} & \textbf{85.17} & \textbf{95.25} & \textbf{68.25} & \textbf{91.36} & \textbf{98.34} \\
\midrule
\multirow{2}{*}{ResNet18}
& D3   & 1.93  & 34.70 & 68.55 & 1.68  & 53.89 & 85.89 & 0.94  & 43.73 & 87.99 \\
& \framework{} & \textbf{61.23} & \textbf{77.87} & \textbf{88.01} & \textbf{76.10} & \textbf{89.38} & \textbf{96.57} & \textbf{73.53} & \textbf{89.35} & \textbf{96.65} \\
\midrule
\multirow{2}{*}{VGG16}
& D3   & 4.32  & 51.01 & 74.85 & 2.48  & 68.33 & 87.57 & 1.67  & 65.61 & 92.85 \\
& \framework{} & \textbf{58.75} & \textbf{73.70} & \textbf{85.18} & \textbf{78.62} & \textbf{88.77} & \textbf{95.63} & \textbf{70.59} & \textbf{85.53} & \textbf{94.63} \\
\midrule
\multirow{2}{*}{EfficientNet-b4}
& D3   & 1.41  & 21.06 & 51.74 & 3.64  & 40.52 & 72.99 & 0.97  & 26.81 & 66.49 \\
& \framework{} & \textbf{49.03} & \textbf{73.38} & \textbf{87.46} & \textbf{60.25} & \textbf{81.24} & \textbf{93.02} & \textbf{56.74} & \textbf{85.39} & \textbf{95.99} \\
\midrule
\multirow{2}{*}{MobileNet-v3}
& D3   & 1.77  & 31.60 & 62.67 & 1.60  & 50.15 & 80.48 & 1.02  & 41.68 & 82.13 \\
& \framework{} & \textbf{49.81} & \textbf{71.03} & \textbf{86.02} & \textbf{67.15} & \textbf{83.76} & \textbf{94.42} & \textbf{61.07} & \textbf{84.57} & \textbf{95.52} \\
\bottomrule
\end{tabular}
}
\end{table}
\subsection{Effect of the Vision Encoder}
\label{sec:ablation_encoder}

We ablate the frozen vision encoder used for patch token extraction, as summarized in \cref{tab:comparison}, comparing \framework{} against D3 under cross-real threshold transfer across all three benchmarks. \framework{} consistently and substantially outperforms D3 with every tested backbone, spanning transformer (CLIP, XCLIP, DINOv2) and CNN (ResNet18, VGG16, EfficientNet-b4, MobileNet-v3) architectures, with margins peaking at the deployment-critical FPR\,=\,0.1\%. This universal improvement confirms that the gains stem from patch-level spatiotemporal scoring rather than from any particular pretrained representation.

Encoder choice modulates performance but does not alter the conclusion. Transformer backbones with finer spatial granularity yield the strongest results, B/16 variants consistently outperform their B/32 counterparts, consistent with our hypothesis that localized artifacts are better captured at higher patch resolution. Lightweight CNN encoders still deliver large improvements over D3, demonstrating that the detection signal is encoder-agnostic. These results jointly show that replacing D3's global clip-level aggregation with localized TTR and LSMI measurement is the dominant driver of performance, while stronger vision tokenizers further amplify this advantage under strict real-only calibration.

\section{Conclusion}
\label{sec:conclusion}

We presented \framework{}, a training-free detector for AI-generated and partially edited videos that measures localized manipulation evidence on patch tokens from a frozen vision encoder through two complementary signals---Two-step Temporal Roughness (TTR) and Local Spatial Motion Incoherence (LSMI)---fused via gamma correction without any learned parameters. Together with a service-aligned evaluation protocol based on Fake Recall at fixed FPR with real-only threshold calibration and cross-real threshold transfer, we evaluated \framework{} across three benchmarks spanning partial edits (FakeParts), near-domain generators (GenVideo), and out-of-distribution 2024--2025 generators (ViF-Bench). \framework{} consistently achieved the highest Fake Recall at FPR targets as strict as $0.1\%$, substantially outperforming both supervised and training-free baselines while remaining robust to post-processing with negligible overhead. These results demonstrate that localized spatiotemporal incoherence on frozen patch tokens provides a strong and generalizable signal for reliable video manipulation detection under practical deployment constraints. We advocate reporting service-aligned operating points, such as Fake Recall at $\alpha=0.1\%$ with real-only threshold calibration and cross-real threshold transfer, as a key metric for real-world AI-generated video detection.

\section*{Acknowledgements}
This work was supported by Institute of Information \& Communications Technology Planning \& Evaluation (IITP) grant funded by the Korea government (MSIT) (No.~RS-2019-II191906, Artificial Intelligence Graduate School Program (POSTECH)).

%
%
\bibliographystyle{splncs04}
\bibliography{main}

\clearpage
\section*{Appendix}
\setcounter{section}{0}
\renewcommand{\thesection}{\Alph{section}}
\renewcommand{\theHsection}{appendix.\Alph{section}}

\section{Benchmark Statistics}
\label{sec:benchmark_statistics}

This section provides detailed category and generator breakdowns for each test benchmark used in our experiments.

\paragraph{FakeParts.}

FakeParts~\cite{liu2025fakeparts} contains 7{,}252 real videos from the ROVI~\cite{wu2024languagedriven} dataset and 64{,}253 fake videos spanning six partial-edit and two full-generation categories: T2V~\cite{hacohen2024ltxvideorealtimevideolatent, kong2025hunyuanvideosystematicframeworklarge, genmo2024mochi, yang2025cogvideox, brooks2024video, vandenoord2024veo2, wan2025wana, zheng2024opensorademocratizingefficientvideo, zhou2024allegroopenblackbox} (16{,}119), Inpainting~\cite{li2025diffueraserdiffusionmodelvideo, wu2024languagedriven, Zhou_2023_ICCV} (13{,}577), Interpolation~\cite{wang2025framer} (10{,}000), Outpainting~\cite{Wang_2025_CVPR} (7{,}275), Style Change~\cite{kara2024rave, ku2024anyvv} (5{,}266), Faceswap~\cite{Deng_2019_CVPR} (5{,}036), TI2V~\cite{wan2025wana, zheng2024opensorademocratizingefficientvideo} (3{,}980), and Extrapolation~\cite{nvidia2025cosmosworldfoundationmodel} (3{,}000).

\paragraph{GenVideo.}

GenVideo~\cite{chen2024demamba} pairs 10{,}000 real videos from MSR-VTT~\cite{xu2016msrvtt} with 8{,}588 fake videos from ten earlier text-to-video generators: Lavie~\cite{wang2025lavie} (1{,}400), Crafter~\cite{chen2023videocrafter1opendiffusionmodels} (1{,}400), Gen2~\cite{Esser_2023_ICCV} (1{,}380), WildScrape~\cite{chen2024demamba} (926), ModelScope~\cite{wang2023modelscopetexttovideotechnicalreport} (700), MorphStudio~\cite{morph2023morphstudio} (700), MoonValley~\cite{moonvalley2022moonvalley} (626), HotShot~\cite{Mullan_Hotshot-XL_2023} (700), Show\_1~\cite{zhang2025show1} (700), and Sora~\cite{brooks2024video} (56).

\paragraph{ViF-Bench.}

ViF-Bench~\cite{li2025skyra} targets newer 2024--2025 generation systems and contains 3{,}650 fake videos. It spans fourteen T2V generators and five I2V generators: CogVideoX1.5-T~\cite{yang2025cogvideox} (200), HunyuanVideo~\cite{kong2025hunyuanvideosystematicframeworklarge} (199), HunyuanVideo-I2V~\cite{kong2025hunyuanvideosystematicframeworklarge} (200), LTX-Video (I2V)~\cite{hacohen2024ltxvideorealtimevideolatent} (200), LTX-Video (T2V)~\cite{hacohen2024ltxvideorealtimevideolatent} (199), SkyReels-V2 (T2V)~\cite{chen2025skyreelsv2infinitelengthfilmgenerative} (200), SkyReels-V2 (I2V)~\cite{chen2025skyreelsv2infinitelengthfilmgenerative} (199), Wan2.1-T2V-1.3B~\cite{wan2025wana} (200), VACE-1.3B-T~\cite{Jiang_2025_ICCV} (200), Wan2.2-I2V-A14B~\cite{wan2025wana} (200), Wan2.2-T2V-A14B~\cite{wan2025wana} (200), Wan2.2-TI2V-5B (I2V)~\cite{wan2025wana} (200), Wan2.2-TI2V-5B (T2V)~\cite{wan2025wana} (200), Gen4-turbo~\cite{runway} (154), Hailuo-02~\cite{hailuo} (172), Kling-V1~\cite{klinga} (176), Pika-V2~\cite{pika} (186), Pixverse-V4-5~\cite{pixverse} (183), and Sora-2~\cite{openai2025sora} (182).
\section{ViF-Bench at FPR $=1\%$ and $5\%$}
\label{sec:vif_bench}

\begin{table}[t]
\centering
\caption{Fake Recall (\%) on ViF-Bench under cross-real threshold transfer at FPR $= 1\%$. Methods are grouped by horizontal lines. Overall denotes the simple average across all categories. Best results are in \textbf{bold}.}
\label{tab:vif_results_fpr1}
\renewcommand{\arraystretch}{1.0}

\resizebox{\textwidth}{!}{%
\begin{tabular}{lcccccccccccccc|ccccc|c}
\toprule
\multirow{2}{*}{Method} & \multicolumn{14}{c}{T2V Models} & \multicolumn{5}{c}{I2V Models} & \multirow{2}{*}{Overall} \\
\cmidrule(lr){2-15} \cmidrule(lr){16-20}
 & \rotatebox{90}{CogX1.5} & \rotatebox{90}{Hunyuan} & \rotatebox{90}{LTX-T} & \rotatebox{90}{SkyV2} & \rotatebox{90}{Wan2.1} & \rotatebox{90}{Wan2.1V} & \rotatebox{90}{Wan2.2} & \rotatebox{90}{Wan2.2TI-T} & \rotatebox{90}{Gen4} & \rotatebox{90}{Hailuo-02} & \rotatebox{90}{Kling-V1} & \rotatebox{90}{Pika-V2} & \rotatebox{90}{Pix4.5} & \rotatebox{90}{Sora-2} & \rotatebox{90}{Hunyuan-I} & \rotatebox{90}{LTX-I} & \rotatebox{90}{SkyV2-I} & \rotatebox{90}{Wan2.2-I} & \rotatebox{90}{Wan2.2TI-I} &  \\
\midrule
NPR & 0.99 & 1.96 & 0.97 & 0.00 & 0.00 & 0.00 & 0.00 & 0.00 & 1.28 & 2.24 & 4.25 & 6.69 & 2.08 & 0.00 & 0.00 & 1.89 & 0.97 & 0.00 & 0.00 & 1.23 \\
STIL & 2.95 & 3.79 & 15.77 & 5.27 & 5.20 & 1.01 & 7.07 & 3.25 & 1.32 & 4.21 & 2.14 & 1.30 & 3.55 & 1.67 & 2.69 & 1.90 & 1.36 & 1.90 & 0.68 & 3.53 \\
FID & 39.25 & 27.53 & 15.07 & 34.79 & 29.20 & 14.70 & 17.61 & 6.06 & 18.95 & 31.97 & 16.92 & 57.79 & 24.86 & 6.31 & 6.33 & 12.80 & 10.31 & 14.54 & 8.79 & 20.73 \\
\midrule
MINTIME & 47.85 & 42.25 & 42.86 & 47.25 & 57.36 & 12.48 & 44.96 & 20.38 & 35.61 & 58.45 & 54.97 & 59.34 & 65.82 & 20.53 & 0.00 & 16.53 & 12.70 & 13.34 & 11.22 & 34.94 \\
FTCN & 42.77 & 47.20 & 54.46 & 53.56 & 42.80 & 16.46 & 45.86 & 11.76 & 30.83 & 67.01 & 48.61 & 61.03 & 66.48 & 12.59 & 0.51 & 17.31 & 14.76 & 16.32 & 9.88 & 34.75 \\
TALL & 1.73 & 0.89 & 0.68 & 1.01 & 1.21 & 2.13 & 0.89 & 0.00 & 2.96 & 0.00 & 0.00 & 0.00 & 0.55 & 0.84 & 0.95 & 3.65 & 0.82 & 2.50 & 0.93 & 1.14 \\
XCLIP & 4.79 & 12.67 & 2.26 & 16.89 & 6.81 & 4.04 & 14.13 & 3.54 & 6.09 & 14.32 & 2.29 & 3.78 & 7.14 & 3.89 & 0.51 & 4.55 & 2.77 & 4.04 & 0.51 & 6.05 \\
AIGVDet & 36.24 & 50.50 & 58.37 & 70.71 & 46.01 & 36.55 & 55.07 & 29.80 & 24.70 & 66.69 & 33.88 & 52.47 & 63.52 & 32.36 & 2.64 & 21.89 & 18.27 & 24.57 & 13.93 & 38.85 \\
DeMamba & 13.30 & 23.91 & 6.04 & 29.03 & 20.83 & 8.06 & 23.53 & 10.33 & 14.76 & 37.59 & 16.42 & 9.66 & 28.04 & 12.70 & 1.68 & 8.29 & 8.23 & 9.34 & 5.66 & 15.13 \\
ReStraV & 67.33 & 81.69 & 73.31 & 81.79 & 85.55 & 74.93 & 82.84 & 68.94 & 72.63 & 82.48 & 90.28 & 84.73 & 75.87 & 73.53 & 69.69 & 75.08 & 83.09 & 84.70 & 64.38 & 77.52 \\
\midrule
D3 & 50.72 & 39.62 & 78.53 & 58.02 & 43.63 & 34.65 & 53.96 & 34.28 & 40.88 & 48.23 & 70.41 & 34.86 & 62.47 & 46.13 & 44.36 & 37.75 & 44.96 & 38.80 & 26.38 & 46.77 \\
\framework{} & \textbf{89.90} & \textbf{94.15} & \textbf{88.73} & \textbf{95.39} & \textbf{98.22} & \textbf{98.99} & \textbf{94.38} & \textbf{99.75} & \textbf{96.03} & \textbf{89.69} & \textbf{99.71} & \textbf{99.46} & \textbf{94.62} & \textbf{98.05} & \textbf{82.07} & \textbf{94.68} & \textbf{95.41} & \textbf{97.22} & \textbf{100.00} & \textbf{95.08} \\
\bottomrule
\end{tabular}%
}
\end{table}

\begin{table}[t]
\centering
\caption{Fake Recall (\%) on ViF-Bench under cross-real threshold transfer at FPR $= 5\%$.}
\label{tab:vif_results_fpr5}
\renewcommand{\arraystretch}{1.0}

\resizebox{\textwidth}{!}{%
\begin{tabular}{lcccccccccccccc|ccccc|c}
\toprule
\multirow{2}{*}{Method} & \multicolumn{14}{c}{T2V Models} & \multicolumn{5}{c}{I2V Models} & \multirow{2}{*}{Overall} \\
\cmidrule(lr){2-15} \cmidrule(lr){16-20}
 & \rotatebox{90}{CogX1.5} & \rotatebox{90}{Hunyuan} & \rotatebox{90}{LTX-T} & \rotatebox{90}{SkyV2} & \rotatebox{90}{Wan2.1} & \rotatebox{90}{Wan2.1V} & \rotatebox{90}{Wan2.2} & \rotatebox{90}{Wan2.2TI-T} & \rotatebox{90}{Gen4} & \rotatebox{90}{Hailuo-02} & \rotatebox{90}{Kling-V1} & \rotatebox{90}{Pika-V2} & \rotatebox{90}{Pix4.5} & \rotatebox{90}{Sora-2} & \rotatebox{90}{Hunyuan-I} & \rotatebox{90}{LTX-I} & \rotatebox{90}{SkyV2-I} & \rotatebox{90}{Wan2.2-I} & \rotatebox{90}{Wan2.2TI-I} &  \\
\midrule
NPR & 7.32 & 15.54 & 1.98 & 9.47 & 8.66 & 0.97 & 0.00 & 0.00 & 15.79 & 15.90 & 16.87 & 17.20 & 9.55 & 0.00 & 0.87 & 6.18 & 6.13 & 0.98 & 1.52 & 7.10 \\
STIL & 11.96 & 12.94 & 32.14 & 18.15 & 18.20 & 6.22 & 15.08 & 10.39 & 4.91 & 10.75 & 11.08 & 9.64 & 13.65 & 8.96 & 7.54 & 5.00 & 4.00 & 6.02 & 3.25 & 11.05 \\
FID & 58.67 & 51.40 & 40.41 & 55.88 & 54.94 & 23.77 & 42.14 & 15.15 & 27.13 & 57.97 & 37.53 & 81.22 & 50.07 & 16.98 & 14.87 & 22.38 & 20.64 & 20.57 & 14.92 & 37.19 \\
\midrule
MINTIME & 65.66 & 69.79 & 69.04 & 71.46 & 85.35 & 37.62 & 70.96 & 46.97 & 66.45 & 83.62 & 84.57 & 84.86 & 90.88 & 42.50 & 2.02 & 41.65 & 33.24 & 33.84 & 25.76 & 58.22 \\
FTCN & 65.72 & 64.27 & 69.21 & 68.56 & 67.15 & 31.91 & 63.07 & 25.20 & 46.48 & 80.87 & 66.51 & 80.77 & 80.08 & 30.19 & 2.42 & 27.85 & 23.69 & 23.35 & 16.43 & 49.14 \\
TALL & 2.80 & 6.25 & 5.29 & 3.91 & 3.96 & 9.31 & 7.17 & 1.57 & 10.66 & 3.64 & 2.63 & 0.99 & 4.50 & 3.33 & 12.99 & 11.45 & 6.49 & 10.83 & 8.52 & 6.12 \\
XCLIP & 18.77 & 33.44 & 22.29 & 39.23 & 21.67 & 12.64 & 36.30 & 11.94 & 15.68 & 36.17 & 19.85 & 13.42 & 32.03 & 14.67 & 2.25 & 11.22 & 9.84 & 12.10 & 7.04 & 19.50 \\
AIGVDet & 47.35 & 62.66 & 70.92 & 78.39 & 61.51 & 43.32 & 65.49 & 37.44 & 31.67 & 80.80 & 46.09 & 64.17 & 71.85 & 39.35 & 3.56 & 29.51 & 27.14 & 30.63 & 21.12 & 48.05 \\
DeMamba & 35.93 & 47.47 & 38.35 & 56.48 & 46.75 & 23.89 & 46.26 & 29.36 & 28.21 & 64.70 & 40.42 & 31.81 & 60.06 & 31.71 & 8.39 & 22.55 & 19.86 & 21.76 & 19.76 & 35.46 \\
ReStraV & 89.07 & 95.21 & 89.95 & 94.68 & 97.67 & 95.24 & 95.24 & 90.30 & 90.76 & 95.73 & 98.84 & 96.93 & 92.90 & 93.17 & 90.27 & 94.68 & 94.08 & 95.51 & 85.88 & 93.48 \\
\midrule
D3 & 84.08 & 86.95 & \textbf{97.39} & 95.59 & 86.23 & 81.83 & 91.92 & 76.74 & 85.78 & 91.44 & 96.49 & 83.77 & 92.44 & 86.76 & 86.29 & 89.27 & 89.11 & 88.48 & 77.17 & 87.78 \\
\framework{} & \textbf{91.92} & \textbf{97.97} & 96.95 & \textbf{98.99} & \textbf{99.49} & \textbf{99.49} & \textbf{100.00} & \textbf{100.00} & \textbf{98.68} & \textbf{97.66} & \textbf{100.00} & \textbf{100.00} & \textbf{100.00} & \textbf{99.44} & \textbf{95.71} & \textbf{96.97} & \textbf{98.98} & \textbf{98.48} & \textbf{100.00} & \textbf{98.46} \\
\bottomrule
\end{tabular}%
}
\end{table}

As shown in \cref{tab:vif_results_fpr1,tab:vif_results_fpr5}, under cross-real threshold transfer on ViF-Bench, \framework{} remains highly effective as the operating constraint relaxes, improving from 84.52\% at FPR\,=\,0.1\% to 95.08\% at 1\% and 98.46\% at 5\% overall. At FPR\,=\,1\%, \framework{} achieves uniformly high recall across both T2V and I2V models, substantially exceeding the strongest baselines (\eg, D3: 46.77\%, AIGVDet: 38.85\%). At FPR\,=\,5\%, performance is near-saturated across almost all generators, while baselines improve but still trail markedly overall (D3: 87.78\%, AIGVDet: 48.05\%, MINTIME: 58.22\%); the only notable exception is LTX-Video (T2V) where D3 is marginally higher (97.39\% vs. 96.95\%). These results reinforce the paper's conclusion that patch-level spatiotemporal incoherence cues (TTR/LSMI) provide strong, calibration-stable separation on modern, unseen generators without retraining, and that \framework{} reaches high recall quickly as the allowed FPR increases.
\section{Effect and Interpretation of the TTR Step Setting}
\label{sec:ttr_analaysis}

\subsection{Effect of TTR Step Setting}
\label{sec:ablation_ttr_steps}
\begin{table}[tb]
\centering
\caption{FakeParts Overall Fake Recall (\%) at fixed FPR (0.1/1/5\%) for different TTR step settings.}
\label{tab:ttr-result}

\begin{tabular}{lccc}
\toprule
\multirow{2}{*}{Step} & \multicolumn{3}{c}{Overall} \\
\cmidrule(lr){2-4}
 & 0.1\% & 1\% & 5\% \\
\midrule
One-step   & 9.54 & 13.54 & 20.66 \\
Two-step   & \textbf{74.45} & \textbf{83.58} & \textbf{89.70} \\
Three-step & 61.64 & 74.92 & 84.25 \\
Four-step  & 46.01 & 58.99 & 70.94 \\
\bottomrule
\end{tabular}

\end{table}

We ablate the maximum temporal step size used by TTR, which controls how many temporal scales contribute to the roughness estimate. For each step size $k$, we first compute a normalized temporal curve length

\begin{equation}
    \hat{L}_k(n) = \sum_{t=1}^{T-k}\|p_{t+k,n}-p_{t,n}\|_2,
    \qquad
    L_k(n) = \left(\frac{\hat{L}_k(n)}{k}\right)\cdot \frac{T-1}{T-k},
\end{equation}
which requires $T \ge k+1$. We then derive the final TTR score depending on the maximum temporal step size.

With one-step ($K=1$), there is no cross-scale comparison, and we use the single-scale proxy
\begin{equation}
    \mathrm{TTR}_1(n) = \log_2(L_1(n)+\epsilon).
\end{equation}

With two-step ($K=2$), TTR becomes the log-ratio between 1-step and 2-step variations:
\begin{equation}
    \mathrm{TTR}_2(n) = \log_2(L_1(n)+\epsilon)-\log_2(L_2(n)+\epsilon).
\end{equation}

When the maximum temporal step size is $K=3$ or $K=4$, we compute $L_k(n)$ for all integer step sizes $k=1,\dots,K$, fit a least-squares line to the points
\begin{equation}
    \left\{(\log_2(1/k),\, \log_2(L_k(n)+\epsilon))\right\}_{k=1}^{K},
\end{equation}
and use the fitted slope as the patch-wise score $\mathrm{TTR}_K(n)$. Finally, we obtain a clip-level statistic by averaging across patches:
\begin{equation}
    \mathbf{TTR}_K = \frac{1}{N}\sum_{n=1}^{N}\mathrm{TTR}_K(n).
\end{equation}

As shown in \cref{tab:ttr-result}, two-step is optimal for FakeParts Overall under our service-aligned operating points. In contrast, one-step performs poorly, indicating that absolute short-term volatility alone is insufficient under real-only threshold calibration. Interestingly, adding more scales reduces performance, suggesting that the most discriminative cue in this setting is the inconsistency between 1-step and 2-step evolution; incorporating longer steps tends to dilute this local inconsistency and yields weaker separation at ultra-low FPR.

\subsection{Patch-wise Qualitative Analysis of Two-Step Separation}
\label{sec:qual_ttr_steps}
\begin{figure}[t]
  \centering
    \includegraphics[width=\linewidth]{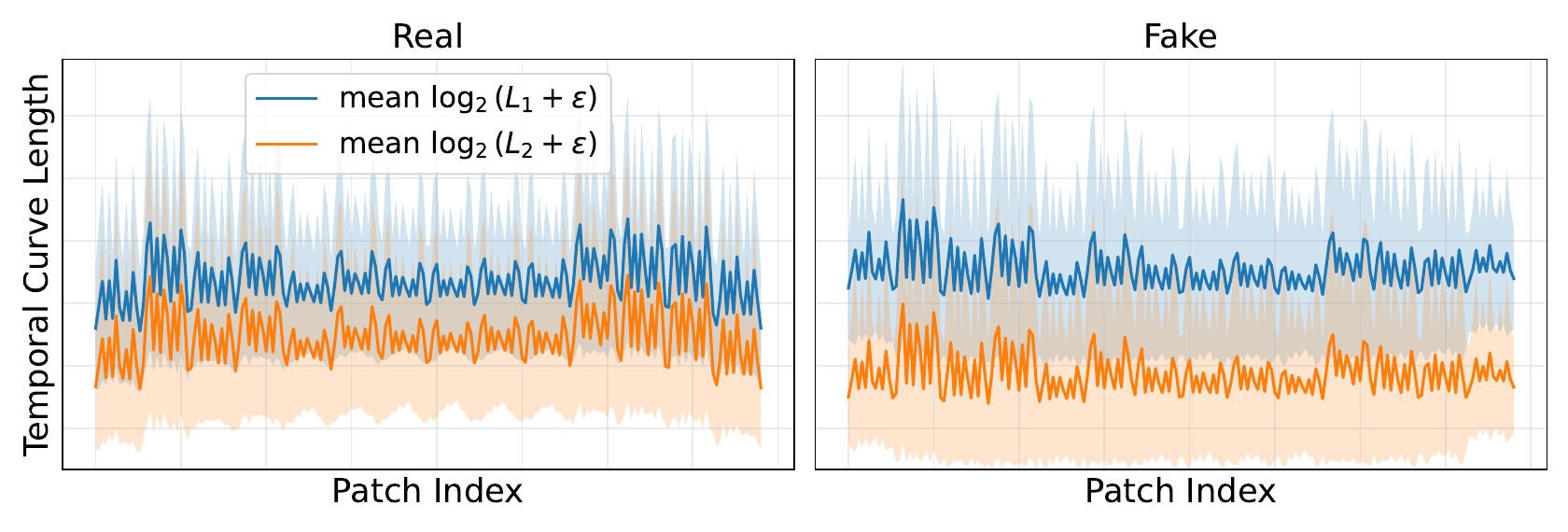}
    \caption{Patch-wise visualization of the two TTR components on FakeParts (left: real videos from ROVI; right: fake videos from Style Change). For each patch index $n$, we plot the mean $\log_2(L_1(n)+\epsilon)$ and mean $\log_2(L_2(n)+\epsilon)$, with shaded bands denoting one standard deviation across videos.}
  \label{fig:ttr-qual}
\end{figure}

A patch-wise view of the two quantities used to compute TTR is shown in \cref{fig:ttr-qual}. For each patch index $n$, we plot the mean $\log_2(L_1(n)+\epsilon)$ and mean $\log_2(L_2(n)+\epsilon)$ separately for real and fake videos, with shaded regions indicating standard deviation across videos. Since
\[
\mathrm{TTR}(n)=\log_2(L_1(n)+\epsilon)-\log_2(L_2(n)+\epsilon),
\]
the vertical gap between the two curves at a given patch index directly reflects the average per-patch roughness signal.

Two observations are clear. First, for real videos, the two curves remain relatively close across most patch indices. This indicates that the accumulated 1-step path length and the normalized 2-step chord-length proxy are well matched, which is consistent with smooth and self-consistent temporal evolution in feature space. Second, for fake videos, the gap between the two curves is larger over a broad range of patch indices. In other words, adjacent-frame changes accumulate more strongly than their 2-step counterpart, which is consistent with jagged temporal trajectories caused by generation or editing artifacts such as flicker, texture refresh, or local temporal misalignment.
\section{Stratified Bootstrap Confidence Intervals}
\label{sec:supp_bootstrap_ci}

We estimate uncertainty for FakeParts Overall Fake Recall@0.1\% under the same cross-real threshold transfer protocol used in the main paper. The bootstrap unit is a video. Detector scores are computed once, and bootstrap resampling is applied to the resulting video-level scores.

Let $\mathcal{R}_{M}$ and $\mathcal{R}_{R}$ denote the MSR-VTT and ROVI real calibration sets, respectively, and let $\mathcal{F}_{c}$ denote the FakeParts fake videos in manipulation category $c \in \mathcal{C}$. We use the eight FakeParts categories in $\mathcal{C}$. For each bootstrap replicate $b$, we independently resample $\mathcal{R}_{M}$ and $\mathcal{R}_{R}$ with replacement to their original sizes. These resampled real sets are used only for threshold estimation. We also resample fake videos with replacement within each category, preserving the category-wise Overall aggregation used in FakeParts.

For target FPR $\alpha=0.1\%$, we compute thresholds from the resampled real calibration scores as
\begin{equation}
\tau_{M,\alpha}^{(b)}
=
Q^{\mathrm{higher}}_{1-\alpha}
\left(\{s(x):x\in\mathcal{R}_{M}^{(b)}\}\right),
\qquad
\tau_{R,\alpha}^{(b)}
=
Q^{\mathrm{higher}}_{1-\alpha}
\left(\{s(x):x\in\mathcal{R}_{R}^{(b)}\}\right),
\end{equation}
where $Q^{\mathrm{higher}}$ denotes the empirical quantile. A video is classified as fake when
$s(x)\ge\tau$.

For each category $c$, we evaluate recall on the bootstrapped fake set
$\mathcal{F}_{c}^{(b)}$ under both real calibration thresholds:
\begin{equation}
\rho_{c,M}^{(b)}
=
\frac{1}{|\mathcal{F}_{c}^{(b)}|}
\sum_{x\in\mathcal{F}_{c}^{(b)}}
\mathbf{1}[s(x)\ge\tau_{M,\alpha}^{(b)}],
\qquad
\rho_{c,R}^{(b)}
=
\frac{1}{|\mathcal{F}_{c}^{(b)}|}
\sum_{x\in\mathcal{F}_{c}^{(b)}}
\mathbf{1}[s(x)\ge\tau_{R,\alpha}^{(b)}].
\end{equation}
The category-level cross-real transfer score is the harmonic mean
\begin{equation}
h_{c}^{(b)}
=
\operatorname{HM}\left(\rho_{c,M}^{(b)},\rho_{c,R}^{(b)}\right)
=
\frac{
2\rho_{c,M}^{(b)}\rho_{c,R}^{(b)}
}{
\rho_{c,M}^{(b)}+\rho_{c,R}^{(b)}
},
\end{equation}
with $h_c^{(b)}=0$ if either recall is zero. The FakeParts Overall score for replicate $b$ is then the arithmetic mean over categories:
\begin{equation}
\rho_{\mathrm{Overall}}^{(b)}
=
\frac{1}{|\mathcal{C}|}
\sum_{c\in\mathcal{C}} h_c^{(b)}.
\end{equation}
The point estimate reported in the main paper is computed using the original, non-resampled real and fake sets with the same aggregation rule. We use $B=10{,}000$ bootstrap replicates and report the percentile 95\% confidence interval given by the 2.5th and 97.5th percentiles of
$\{\rho_{\mathrm{Overall}}^{(b)}\}_{b=1}^{B}$.

\begin{table}[t]
\centering
\caption{
Stratified-bootstrap 95\% confidence intervals for FakeParts Overall Fake Recall@0.1\% under cross-real threshold transfer. All values are percentages.
}
\label{tab:supp_bootstrap_ci}
\begin{tabular}{lcc}
\toprule
Method & Fake Recall@0.1\% & 95\% CI \\
\midrule
D3 & 2.28 & $[0.53,\,5.43]$ \\
\framework{} & 74.45 & $[72.42,\,76.14]$ \\
\bottomrule
\end{tabular}
\end{table}

As shown in \cref{tab:supp_bootstrap_ci}, the confidence intervals for D3 and \framework{} do not overlap. This indicates that the gain at the deployment-critical FPR\,=\,0.1\% operating point is stable under resampling of both the real calibration data and the category-stratified FakeParts fake videos.

\section{Additional Sensitivity to the Gamma Exponent}
\label{sec:supp_gamma_sensitivity}
\begin{figure}[t]
    \centering
    \includegraphics[width=\linewidth]{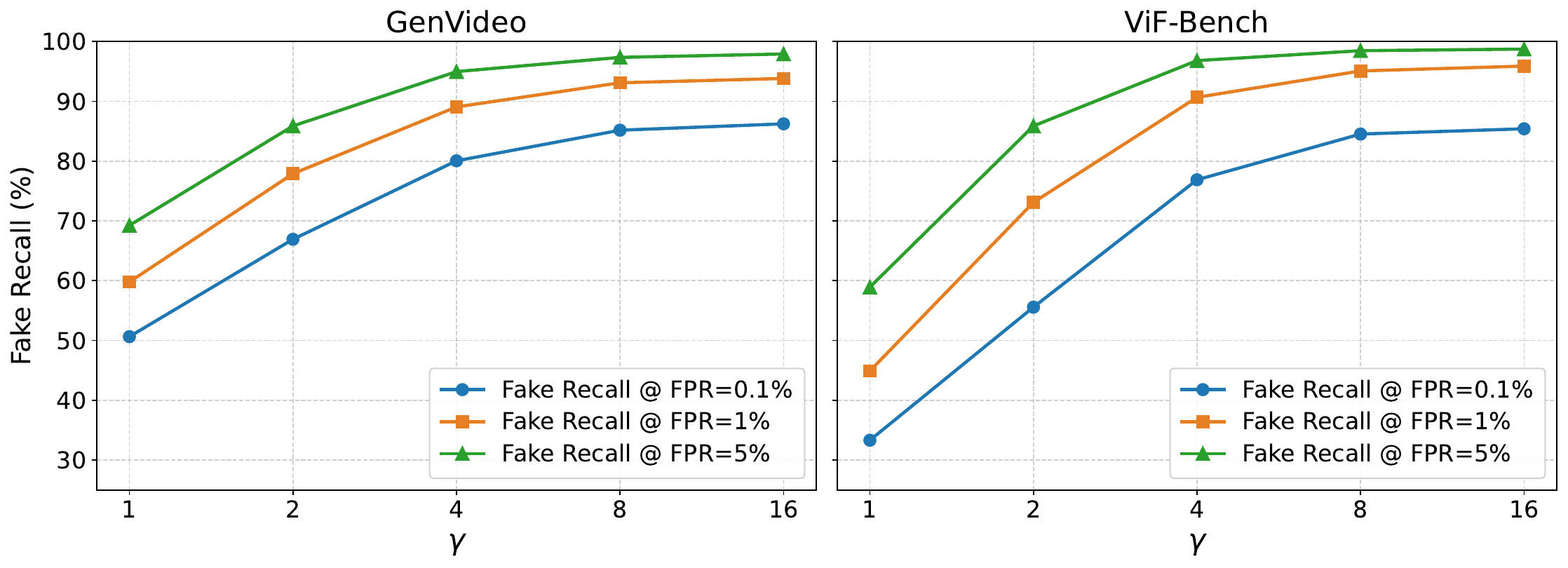}
    \caption{Additional sensitivity to the gamma exponent on GenVideo and ViF-Bench.}
    \label{fig:supp_gamma_vif}
\end{figure}

To verify that the choice of $\gamma=8$ is not specific to the FakeParts ablation in the main paper, we repeat the $\gamma$ sweep on GenVideo and ViF-Bench using the same evaluation protocol. We vary $\gamma\in\{1,2,4,8,16\}$ in $s(x)=\mathbf{TTR}^{\gamma}\cdot\mathbf{LSMI}$ while keeping all other settings fixed, including the XCLIP-B/16 encoder, real-only threshold calibration, and cross-real threshold transfer. As shown in \cref{fig:supp_gamma_vif}, Fake Recall consistently improves from $\gamma=1$ to $\gamma=8$ across FPR targets and then saturates, with little additional gain at $\gamma=16$. These results match the FakeParts trend and support using a single fixed value, $\gamma=8$, for all datasets and FPR targets without per-benchmark tuning.

\section{AUROC Results and Analysis}
\label{sec:auroc_analysis}

\begin{table}[p]
\centering
\caption{Cross-real AUROC (\%, harmonic mean across real domains) on FakeParts. Methods are grouped by horizontal lines. Overall denotes the simple average across all categories. Best results are in \textbf{bold}.}
\label{tab:supp_fakeparts_auroc}
\renewcommand{\arraystretch}{1.0}
\setlength{\tabcolsep}{4pt}

\resizebox{\textwidth}{!}{%
\begin{tabular}{lccccccccc}
\toprule
Method & Extrapolation & Faceswap & Inpainting & Interpolation & Outpainting & Style Change & T2V & TI2V & Overall \\
\midrule
NPR & 60.70 & 38.22 & 65.08 & 50.45 & 31.39 & 46.84 & 71.99 & 78.06 & 55.34 \\ 
STIL & 37.84 & 45.04 & 56.23 & 43.05 & 39.45 & 64.09 & 78.89 & 76.59 & 55.15 \\
FID & 64.62 & 82.85 & \textbf{78.55} & 80.19 & 84.19 & 89.49 & 87.99 & 92.39 & 82.53 \\
\midrule
MINTIME & 33.90 & 28.96 & 37.54 & 73.68 & 42.81 & 83.33 & 77.41 & 90.91 & 58.57 \\
FTCN & 40.66 & 42.16 & 38.66 & 60.00 & 34.40 & 89.03 & 81.09 & 94.40 & 60.05 \\
TALL & 49.44 & 39.90 & 60.07 & 58.51 & 55.70 & 54.62 & 53.77 & 44.71 & 52.09 \\
XCLIP & 28.83 & 39.77 & 52.40 & 43.58 & 28.83 & 70.27 & 73.70 & 89.90 & 53.41 \\
AIGVDet & 56.52 & 59.76 & 70.82 & 60.63 & 53.59 & 34.32 & 77.10 & 94.87 & 63.45 \\
DeMamba & 24.91 & 39.46 & 46.05 & 59.75 & 51.54 & 77.70 & 73.37 & 86.59 & 57.42 \\
ReStraV & 96.43 & 96.18 & 50.40 & 92.64 & \textbf{99.92} & 98.02 & 98.23 & 99.46 & 91.41 \\
\midrule
D3 & 93.79 & 97.47 & 62.48 & 92.70 & 98.72 & 96.26 & \textbf{98.64} & 98.00 & 92.26 \\
\framework{} & \textbf{99.97} & \textbf{99.25} & 74.30 & \textbf{99.22} & 99.89 & \textbf{99.98} & 96.73 & \textbf{99.94} & \textbf{96.16} \\
\bottomrule
\end{tabular}%
}
\end{table}

\begin{table}[p]
\centering
\caption{Cross-real AUROC (\%, harmonic mean across real domains) on GenVideo.}  
\label{tab:supp_genvideo_auroc}
\renewcommand{\arraystretch}{1.0}
\setlength{\tabcolsep}{3pt}

\resizebox{\textwidth}{!}{%
\begin{tabular}{lccccccccccc}
\toprule
Method & Crafter & Gen2 & HotShot & Lavie & ModelScope & MoonValley & MorphStudio & Show\_1 & Sora & WildScrape & Overall \\
\midrule
NPR & 81.08 & 74.64 & 81.32 & 71.83 & 75.37 & 81.85 & 76.58 & 51.75 & 64.40 & 66.81 & 72.56 \\
STIL & 91.77 & 90.83 & 85.44 & 83.88 & 85.82 & 94.14 & 89.04 & 85.21 & 83.02 & 76.48 & 86.56 \\
FID & 99.43 & 95.23 & 99.31 & 97.80 & 97.53 & 99.56 & 98.13 & 97.53 & 74.39 & 87.87 & 94.68 \\
\midrule
MINTIME & 98.22 & 97.40 & 81.24 & 97.41 & 89.86 & 96.77 & 97.77 & 94.47 & 70.09 & 91.00 & 91.42 \\
FTCN & 99.23 & 97.89 & 93.58 & 97.14 & 92.25 & 98.29 & 98.81 & 94.59 & 71.78 & 91.07 & 93.46 \\
TALL & 52.23 & 61.69 & 58.39 & 44.78 & 56.75 & 52.62 & 55.80 & 62.73 & 56.10 & 50.58 & 55.17 \\
XCLIP & 94.96 & 94.28 & 86.23 & 85.82 & 91.47 & 94.45 & 94.37 & 87.64 & 85.75 & 82.37 & 89.73 \\
AIGVDet & 99.44 & 97.98 & 5.90 & 83.50 & 97.97 & \textbf{99.60} & 99.02 & 81.17 & 94.57 & 78.08 & 83.72 \\
DeMamba & 96.16 & 97.05 & 85.91 & 89.40 & 87.55 & 87.58 & 94.96 & 94.25 & 73.02 & 80.64 & 88.65 \\
ReStraV & 98.50 & 98.91 & 99.65 & 97.48 & 98.15 & 98.03 & 98.89 & 99.40 & 99.33 & 90.82 & 97.92 \\
\midrule
D3 & 98.67 & \textbf{99.47} & 98.06 & 97.60 & 96.29 & 99.35 & 98.25 & 98.95 & 98.31 & 92.03 & 97.70 \\
\framework{} & \textbf{99.90} & 99.20 & \textbf{100.00} & \textbf{99.70} & \textbf{99.53} & 99.22 & \textbf{99.94} & \textbf{99.99} & \textbf{99.95} & \textbf{92.43} & \textbf{98.99} \\
\bottomrule
\end{tabular}%
}
\end{table}

\begin{table}[p]
\centering
\caption{Cross-real AUROC (\%, harmonic mean across real domains) on ViF-Bench.}
\label{tab:vif_results_auroc}
\renewcommand{\arraystretch}{1.2}

\resizebox{\textwidth}{!}{%
\begin{tabular}{lcccccccccccccc|ccccc|c}
\toprule
\multirow{2}{*}{Method} & \multicolumn{14}{c}{T2V Models} & \multicolumn{5}{c}{I2V Models} & \multirow{2}{*}{Overall} \\
\cmidrule(lr){2-15} \cmidrule(lr){16-20}
 & \rotatebox{90}{CogX1.5} & \rotatebox{90}{Hunyuan} & \rotatebox{90}{LTX-T} & \rotatebox{90}{SkyV2} & \rotatebox{90}{Wan2.1} & \rotatebox{90}{Wan2.1V} & \rotatebox{90}{Wan2.2} & \rotatebox{90}{Wan2.2TI-T} & \rotatebox{90}{Gen4} & \rotatebox{90}{Hailuo-02} & \rotatebox{90}{Kling-V1} & \rotatebox{90}{Pika-V2} & \rotatebox{90}{Pix4.5} & \rotatebox{90}{Sora-2} & \rotatebox{90}{Hunyuan-I} & \rotatebox{90}{LTX-I} & \rotatebox{90}{SkyV2-I} & \rotatebox{90}{Wan2.2-I} & \rotatebox{90}{Wan2.2TI-I} &  \\
\midrule
NPR & 70.27 & 68.26 & 62.41 & 60.76 & 63.39 & 34.79 & 49.65 & 43.15 & 63.21 & 69.89 & 75.35 & 77.97 & 65.11 & 47.83 & 35.92 & 59.52 & 53.74 & 40.37 & 33.70 & 56.59 \\
STIL & 70.19 & 64.06 & 71.00 & 68.90 & 70.06 & 51.71 & 62.84 & 60.92 & 48.56 & 60.56 & 60.42 & 67.13 & 66.16 & 52.00 & 51.81 & 47.55 & 43.90 & 43.95 & 41.52 & 58.07 \\
FID & 84.47 & 83.15 & 81.46 & 82.90 & 85.07 & 56.39 & 79.54 & 58.14 & 59.72 & 86.27 & 77.86 & 94.86 & 83.90 & 62.55 & 64.93 & 58.68 & 51.53 & 50.96 & 43.75 & 70.85 \\
\midrule
MINTIME & 93.84 & 93.86 & 92.67 & 94.54 & 97.19 & 82.96 & 94.29 & 85.92 & 92.30 & 96.90 & 96.60 & 97.08 & 97.97 & 85.77 & 30.39 & 85.57 & 82.25 & 78.74 & 74.16 & 87.00 \\
FTCN & 93.98 & 90.25 & 93.75 & 92.62 & 92.07 & 76.77 & 91.08 & 76.20 & 84.40 & 95.84 & 91.95 & 96.45 & 96.75 & 76.30 & 30.84 & 76.04 & 69.66 & 67.11 & 63.43 & 81.87 \\
TALL & 41.36 & 56.80 & 62.42 & 61.63 & 53.41 & 52.18 & 53.72 & 43.23 & 53.35 & 56.46 & 48.85 & 44.26 & 56.65 & 54.31 & 59.37 & 51.35 & 47.23 & 50.97 & 46.84 & 52.34 \\
XCLIP & 64.25 & 76.32 & 81.15 & 83.13 & 75.50 & 55.99 & 81.15 & 64.12 & 59.42 & 81.10 & 74.20 & 62.99 & 81.99 & 63.21 & 39.86 & 58.17 & 55.40 & 53.06 & 48.35 & 66.28 \\
AIGVDet & 86.17 & 91.28 & 92.41 & 95.50 & 90.68 & 84.94 & 92.50 & 82.94 & 81.08 & 95.64 & 88.11 & 92.24 & 94.12 & 84.81 & 62.59 & 79.79 & 79.59 & 80.41 & 76.52 & 85.86 \\
DeMamba & 81.83 & 87.77 & 85.42 & 91.27 & 89.24 & 75.63 & 87.34 & 81.00 & 78.80 & 92.60 & 87.19 & 82.10 & 91.35 & 80.76 & 46.17 & 77.06 & 72.37 & 73.18 & 70.90 & 80.63 \\
ReStraV & \textbf{97.67} & 98.94 & 98.54 & 99.23 & 99.34 & 98.93 & 99.13 & 97.97 & 98.27 & 99.28 & 99.64 & 99.35 & 98.97 & 98.70 & 96.87 & 98.97 & 99.09 & 99.11 & 98.06 & 98.74 \\
\midrule
D3 & 97.48 & 97.43 & 99.27 & 98.64 & 97.72 & 96.70 & 98.34 & 96.73 & 97.47 & 98.34 & 98.98 & 97.21 & 98.59 & 97.57 & 96.95 & 97.83 & 97.84 & 97.54 & 96.78 & 97.76 \\
\framework{} & 93.56 & \textbf{99.55} & \textbf{99.44} & \textbf{99.79} & \textbf{99.82} & \textbf{99.89} & \textbf{99.83} & \textbf{99.99} & \textbf{99.67} & \textbf{99.32} & \textbf{99.97} & \textbf{99.97} & \textbf{99.81} & \textbf{99.90} & \textbf{98.81} & \textbf{99.09} & \textbf{99.78} & \textbf{99.52} & \textbf{100.00} & \textbf{99.35} \\
\bottomrule
\end{tabular}%
}
\end{table}

For completeness, we report conventional cross-real AUROC (\%, harmonic mean across the two real domains) in \cref{tab:supp_fakeparts_auroc,tab:supp_genvideo_auroc,tab:vif_results_auroc}. Overall, the AUROC results are consistent with the paper's conclusions: \framework{} achieves the best overall performance on all three benchmarks and remains strong across both fully generated and partially edited videos.

On \textbf{FakeParts}, \framework{} attains the highest overall AUROC (96.16\%), outperforming D3 (92.26\%) and all supervised baselines (best: FID, 82.53\%). It ranks first in six of eight manipulation categories, with only Inpainting and T2V led by other methods. Notably, D3 also achieves high AUROC on FakeParts, yet its Fake Recall at ultra-low FPR in the paper is much weaker, indicating that AUROC alone can overstate practical performance for partial-edit detection under strict operating constraints.

On \textbf{GenVideo}, \framework{} reaches a near-saturated overall AUROC of 98.99\%, exceeding D3 (97.70\%) and the best supervised baseline FID (94.68\%). It achieves the highest AUROC on eight of the ten generators, showing consistently strong separation on earlier-generation T2V models.

On \textbf{ViF-Bench}, \framework{} further improves to 99.35\% overall AUROC and achieves the best result on 18 of 19 generators, covering both recent T2V and I2V systems. The gap to supervised detectors remains large (best overall: MINTIME, 87.00\%), confirming strong out-of-distribution generalization without retraining.

Taken together, the AUROC tables corroborate the robustness and generalization of \framework{}. At the same time, the contrast between these strong AUROC values and the much larger gaps observed in Fake Recall at FPR\,=\,0.1\%---especially for D3---reinforces our main claim that deployment-oriented evaluation should prioritize ultra-low-FPR operating points rather than AUROC alone.

\section{Per-Real-Set Results}
\label{sec:supp_per_real_set_results}

This section provides the per-real-set Fake Recall tables referenced in the paper.

\paragraph{FakeParts.}
The FakeParts results with thresholds calibrated using the ROVI and MSR-VTT real sets are reported in \cref{tab:fakeparts_rovi} and \cref{tab:fakeparts_msrvtt}, respectively.

\paragraph{GenVideo.}
The GenVideo results with thresholds calibrated using the ROVI and MSR-VTT real sets are reported in \cref{tab:genvideo_rovi} and \cref{tab:genvideo_msrvtt}, respectively.

\paragraph{ViF-Bench.}
The ViF-Bench results, corresponding to FPR $= 0.1\%, 1\%, 5\%$, are reported in \cref{tab:vif_results_rovi_01,tab:vif_results_rovi_1,tab:vif_results_rovi_5} for thresholds calibrated using the ROVI real set, and in \cref{tab:vif_results_msrvtt_01,tab:vif_results_msrvtt_1,tab:vif_results_msrvtt_5} for the MSR-VTT real set.

\begin{table}[p]
\centering

\caption{Fake Recall (\%) on FakeParts at FPR $\in \{0.1\%,\,1\%,\,5\%\}$, with thresholds calibrated using the ROVI real set only. Methods are grouped by horizontal lines. Overall denotes the simple average across all categories. Best results are in \textbf{bold}.}
\label{tab:fakeparts_rovi}
\renewcommand{\arraystretch}{0.87}
\setlength{\tabcolsep}{4pt}

\resizebox{\textwidth}{!}{%
\begin{tabular}{lccccccccccccccc}
\toprule
\multirow{2}{*}{Method}
 & \multicolumn{3}{c}{Extrapolation} & \multicolumn{3}{c}{Faceswap} & \multicolumn{3}{c}{Inpainting} & \multicolumn{3}{c}{Interpolation} & \multicolumn{3}{c}{Outpainting} \\
\cmidrule(lr){2-4} \cmidrule(lr){5-7} \cmidrule(lr){8-10} \cmidrule(lr){11-13} \cmidrule(lr){14-16}
 & 0.1\% & 1\% & 5\% & 0.1\% & 1\% & 5\% & 0.1\% & 1\% & 5\% & 0.1\% & 1\% & 5\% & 0.1\% & 1\% & 5\% \\
\midrule
NPR & 0.00 & 0.53 & 4.88 & 0.00 & 0.00 & 0.28 & 0.06 & 1.48 & 7.62 & 0.00 & 0.01 & 0.67 & 0.00 & 0.00 & 0.08 \\
STIL & 0.09 & 0.12 & 2.44 & 0.08 & 0.12 & 1.97 & 0.69 & 1.09 & 6.54 & 0.17 & 0.40 & 2.15 & 0.18 & 0.26 & 1.66 \\
FID & 0.00 & 0.37 & 3.03 & 0.08 & 5.01 & 23.62 & 0.18 & 3.70 & 18.03 & 0.98 & 6.55 & 22.79 & 4.03 & 19.97 & 40.82 \\
\midrule
MINTIME & 0.03 & 0.22 & 2.63 & 0.00 & 0.02 & 0.78 & 0.04 & 0.71 & 3.56 & 1.59 & 10.40 & 29.74 & 0.05 & 0.40 & 2.65 \\
FTCN & 0.15 & 1.08 & 3.25 & 0.00 & 0.16 & 1.15 & 0.03 & 0.47 & 2.89 & 0.78 & 4.83 & 13.45 & 0.00 & 0.47 & 1.66 \\
TALL & 0.03 & 0.90 & 3.65 & 0.00 & 0.08 & 0.54 & 0.20 & 1.43 & 5.70 & 0.20 & 1.08 & 5.02 & 0.15 & 0.77 & 3.45 \\
XCLIP & 0.00 & 0.03 & 0.59 & 0.00 & 0.10 & 1.17 & 0.18 & 0.71 & 4.42 & 0.25 & 1.01 & 4.36 & 0.11 & 0.43 & 1.97 \\
AIGVDet & 0.22 & 0.96 & 2.84 & 0.46 & 3.68 & 9.37 & 0.50 & 3.89 & 12.37 & 0.22 & 1.80 & 4.88 & 0.30 & 1.02 & 2.08 \\
DeMamba & 0.00 & 0.03 & 0.28 & 0.02 & 0.12 & 1.29 & 0.04 & 0.66 & 3.46 & 0.00 & 0.78 & 5.89 & 0.08 & 0.82 & 4.30 \\
ReStraV & 86.12 & 97.09 & 99.26 & \textbf{85.71} & \textbf{97.45} & \textbf{99.20} & \textbf{32.82} & \textbf{34.55} & \textbf{38.64} & 72.02 & \textbf{93.80} & \textbf{97.94} & \textbf{99.89} & \textbf{99.95} & \textbf{99.99} \\
\midrule
D3 & 12.77 & 41.02 & 73.01 & 33.35 & 74.11 & 93.29 & 0.94 & 3.88 & 11.48 & 8.83 & 35.27 & 68.56 & 58.24 & 87.16 & 97.21 \\
\framework{} & \textbf{98.61} & \textbf{99.54} & \textbf{99.88} & 71.57 & 88.40 & 96.76 & 6.19 & 19.78 & 34.29 & \textbf{77.04} & 91.11 & 96.83 & 98.20 & 99.51 & 99.86 \\
\bottomrule
\end{tabular}%
}

\resizebox{\textwidth}{!}{%
\begin{tabular}{lccccccccccccc}
\toprule
\multirow{2}{*}{Method} & \multicolumn{3}{c}{Style Change} & \multicolumn{3}{c}{T2V} & \multicolumn{3}{c}{TI2V} & & \multicolumn{3}{c}{Overall} \\
\cmidrule(lr){2-4} \cmidrule(lr){5-7} \cmidrule(lr){8-10} \cmidrule(lr){12-14}
 & 0.1\% & 1\% & 5\% & 0.1\% & 1\% & 5\% & 0.1\% & 1\% & 5\% & & 0.1\% & 1\% & 5\% \\
\midrule
NPR & 0.02 & 0.74 & 2.78 & 0.19 & 2.30 & 11.59 & 0.23 & 2.51 & 13.39 & & 0.06 & 0.95 & 5.16 \\
STIL & 2.19 & 3.37 & 13.77 & 6.72 & 9.60 & 29.16 & 3.52 & 5.53 & 20.60 & & 1.70 & 2.56 & 9.79 \\
FID & 8.29 & 28.35 & 52.12 & 7.38 & 27.72 & 50.95 & 3.44 & 20.65 & 48.24 & & 3.05 & 14.04 & 32.45 \\
\midrule
MINTIME & 2.51 & 18.63 & 42.16 & 1.61 & 11.45 & 31.04 & 3.09 & 22.49 & 56.31 & & 1.11 & 8.04 & 21.11 \\
FTCN & 10.12 & 29.42 & 51.97 & 6.02 & 20.49 & 39.67 & 13.49 & 44.47 & 69.50 & & 3.82 & 12.67 & 22.94 \\
TALL & 0.25 & 1.14 & 4.55 & 0.26 & 1.35 & 4.76 & 0.03 & 0.48 & 2.34 & & 0.14 & 0.90 & 3.75 \\
XCLIP & 2.11 & 8.34 & 24.96 & 5.09 & 14.50 & 32.84 & 11.98 & 29.07 & 56.81 & & 2.46 & 6.77 & 15.89 \\
AIGVDet & 2.17 & 5.87 & 9.49 & 21.84 & 38.46 & 49.66 & 22.09 & 47.04 & 65.33 & & 5.97 & 12.84 & 19.50 \\
DeMamba & 0.80 & 5.92 & 19.64 & 0.98 & 6.59 & 19.10 & 2.14 & 12.29 & 35.43 & & 0.51 & 3.40 & 11.17 \\
ReStraV & 93.03 & 98.86 & 99.87 & \textbf{92.41} & \textbf{99.24} & \textbf{99.84} & \textbf{98.54} & \textbf{99.70} & \textbf{99.95} & & \textbf{82.57} & \textbf{90.08} & \textbf{91.84} \\
\midrule
D3 & 32.79 & 67.40 & 86.87 & 60.19 & 86.04 & 96.85 & 45.48 & 81.81 & 96.03 & & 31.57 & 59.59 & 77.91 \\
\framework{} & \textbf{99.28} & \textbf{99.83} & \textbf{99.92} & 80.51 & 87.07 & 91.89 & 97.06 & 99.05 & 99.87 & & 78.56 & 85.54 & 89.91 \\
\bottomrule
\end{tabular}%
}

\vspace{0.8em}

\captionof{table}{Fake Recall (\%) on FakeParts at FPR $\in \{0.1\%,\,1\%,\,5\%\}$, with thresholds calibrated using the MSR-VTT real set only.}
\label{tab:fakeparts_msrvtt}
\renewcommand{\arraystretch}{0.87}
\setlength{\tabcolsep}{4pt}

\resizebox{\textwidth}{!}{%
\begin{tabular}{lccccccccccccccc}
\toprule
\multirow{2}{*}{Method}
 & \multicolumn{3}{c}{Extrapolation} & \multicolumn{3}{c}{Faceswap} & \multicolumn{3}{c}{Inpainting} & \multicolumn{3}{c}{Interpolation} & \multicolumn{3}{c}{Outpainting} \\
\cmidrule(lr){2-4} \cmidrule(lr){5-7} \cmidrule(lr){8-10} \cmidrule(lr){11-13} \cmidrule(lr){14-16}
 & 0.1\% & 1\% & 5\% & 0.1\% & 1\% & 5\% & 0.1\% & 1\% & 5\% & 0.1\% & 1\% & 5\% & 0.1\% & 1\% & 5\% \\
\midrule
NPR & 6.58 & 16.94 & 30.82 & 0.44 & 1.51 & 5.59 & 9.58 & 19.33 & 34.03 & 1.08 & 5.24 & 14.42 & 0.11 & 0.56 & 2.90 \\
STIL & 0.09 & 0.37 & 3.25 & 0.08 & 0.22 & 3.08 & 0.69 & 1.80 & 8.96 & 0.17 & 0.61 & 3.23 & 0.18 & 0.49 & 2.38 \\
FID & 0.06 & 6.24 & 34.84 & 2.69 & 32.95 & 70.57 & 2.31 & 26.08 & \textbf{64.55} & 4.90 & 31.43 & 68.79 & 15.96 & 47.89 & 74.42 \\
\midrule
MINTIME & 0.06 & 0.68 & 2.63 & 0.00 & 0.06 & 0.78 & 0.14 & 0.96 & 3.51 & 3.43 & 13.29 & 29.37 & 0.11 & 0.66 & 2.56 \\
FTCN & 0.46 & 1.79 & 4.27 & 0.02 & 0.50 & 2.11 & 0.24 & 1.24 & 4.57 & 3.14 & 8.28 & 17.04 & 0.11 & 1.02 & 2.67 \\
TALL & 0.19 & 2.16 & 9.18 & 0.00 & 0.40 & 2.33 & 0.39 & 3.56 & 15.41 & 0.30 & 2.99 & 13.29 & 0.25 & 1.84 & 9.66 \\
XCLIP & 0.00 & 0.03 & 0.87 & 0.00 & 0.16 & 1.63 & 0.05 & 0.87 & 5.55 & 0.14 & 1.05 & 5.15 & 0.08 & 0.47 & 2.45 \\
AIGVDet & 5.60 & 11.44 & 17.13 & 16.14 & 26.80 & 33.39 & \textbf{24.69} & \textbf{40.79} & 49.44 & 8.43 & 13.45 & 17.38 & 3.99 & 7.75 & 11.13 \\
DeMamba & 0.00 & 0.25 & 1.11 & 0.04 & 0.97 & 3.76 & 0.34 & 2.73 & 9.08 & 0.33 & 4.93 & 14.61 & 0.60 & 3.38 & 10.27 \\
ReStraV & 13.45 & 44.98 & 69.24 & 7.02 & 34.28 & 63.77 & 13.46 & 27.63 & 31.21 & 0.09 & 6.25 & 34.19 & 84.32 & 97.91 & 99.59 \\
\midrule
D3 & 0.00 & 10.54 & 49.03 & 0.02 & 28.21 & 81.32 & 0.00 & 0.71 & 5.09 & 0.00 & 6.94 & 44.01 & 1.39 & 53.15 & 91.23 \\
\framework{} & \textbf{97.13} & \textbf{99.20} & \textbf{99.85} & \textbf{51.79} & \textbf{80.40} & \textbf{95.94} & 1.55 & 11.61 & 32.86 & \textbf{60.56} & \textbf{84.29} & \textbf{96.33} & \textbf{94.90} & \textbf{98.96} & \textbf{99.84} \\
\bottomrule
\end{tabular}%
}

\resizebox{\textwidth}{!}{%
\begin{tabular}{lccccccccccccc}
\toprule
\multirow{2}{*}{Method} & \multicolumn{3}{c}{Style Change} & \multicolumn{3}{c}{T2V} & \multicolumn{3}{c}{TI2V} & & \multicolumn{3}{c}{Overall} \\
\cmidrule(lr){2-4} \cmidrule(lr){5-7} \cmidrule(lr){8-10} \cmidrule(lr){12-14}
 & 0.1\% & 1\% & 5\% & 0.1\% & 1\% & 5\% & 0.1\% & 1\% & 5\% & & 0.1\% & 1\% & 5\% \\
\midrule
NPR & 3.39 & 8.61 & 16.96 & 14.31 & 27.63 & 45.25 & 17.06 & 33.19 & 53.59 & & 6.57 & 14.13 & 25.45 \\
STIL & 2.19 & 5.58 & 18.08 & 6.72 & 13.74 & 34.94 & 3.52 & 8.99 & 25.93 & & 1.70 & 3.98 & 12.48 \\
FID & 23.49 & 60.41 & 83.83 & 22.83 & 58.45 & 81.41 & 16.18 & 58.24 & 91.38 & & 11.05 & 40.21 & 71.22 \\
\midrule
MINTIME & 6.02 & 23.39 & 41.74 & 3.41 & 15.02 & 30.81 & 6.76 & 28.84 & 55.90 & & 2.49 & 10.36 & 20.91 \\
FTCN & 22.59 & 40.03 & 59.90 & 14.84 & 29.35 & 46.50 & 34.62 & 57.36 & 76.31 & & 9.50 & 17.45 & 26.67 \\
TALL & 0.38 & 2.99 & 11.56 & 0.42 & 3.14 & 12.14 & 0.13 & 1.36 & 6.16 & & 0.26 & 2.31 & 9.97 \\
XCLIP & 1.24 & 8.99 & 28.35 & 3.18 & 15.27 & 35.85 & 7.91 & 30.45 & 61.13 & & 1.58 & 7.16 & 17.62 \\
AIGVDet & 12.23 & 15.66 & 17.30 & 57.43 & 63.40 & 66.18 & 77.56 & 87.79 & 92.04 & & 25.76 & 33.38 & 38.00 \\
DeMamba & 3.64 & 16.84 & 36.41 & 4.21 & 16.79 & 32.75 & 7.96 & 31.08 & 56.26 & & 2.14 & 9.62 & 20.53 \\
ReStraV & 10.33 & 54.07 & 82.00 & 31.20 & 66.24 & 82.54 & 58.14 & 85.85 & 94.97 & & 27.25 & 52.15 & 69.69 \\
\midrule
D3 & 0.17 & 28.58 & 73.50 & 8.22 & 56.63 & 89.85 & 0.33 & 40.85 & 87.49 & & 1.27 & 28.20 & 65.19 \\
\framework{} & \textbf{98.32} & \textbf{99.60} & \textbf{99.92} & \textbf{73.30} & \textbf{83.67} & \textbf{91.40} & \textbf{93.54} & \textbf{98.17} & \textbf{99.85} & & \textbf{71.39} & \textbf{81.99} & \textbf{89.50} \\
\bottomrule
\end{tabular}%
}
\end{table}
\begin{table}[p]
\centering
\caption{Fake Recall (\%) on GenVideo at FPR $\in \{0.1\%,\,1\%,\,5\%\}$, with thresholds calibrated using the ROVI real set only. Best results are in \textbf{bold}.}  
\label{tab:genvideo_rovi}
\renewcommand{\arraystretch}{1.0}
\setlength{\tabcolsep}{3pt}

\resizebox{\textwidth}{!}{%
\begin{tabular}{lcccccccccccccccccc}
\toprule
\multirow{2}{*}{Method}
 & \multicolumn{3}{c}{Crafter} & \multicolumn{3}{c}{Gen2} & \multicolumn{3}{c}{HotShot} & \multicolumn{3}{c}{Lavie} & \multicolumn{3}{c}{ModelScope} & \multicolumn{3}{c}{MoonValley} \\
\cmidrule(lr){2-4} \cmidrule(lr){5-7} \cmidrule(lr){8-10} \cmidrule(lr){11-13} \cmidrule(lr){14-16} \cmidrule(lr){17-19}
 & 0.1\% & 1\% & 5\% & 0.1\% & 1\% & 5\% & 0.1\% & 1\% & 5\% & 0.1\% & 1\% & 5\% & 0.1\% & 1\% & 5\% & 0.1\% & 1\% & 5\% \\
\midrule
NPR & 0.21 & 4.15 & 19.96 & 0.14 & 1.74 & 8.99 & 0.00 & 0.86 & 12.14 & 0.00 & 0.43 & 5.43 & 0.86 & 5.14 & 15.71 & 0.32 & 2.40 & 15.97 \\
STIL & 22.75 & 28.33 & 56.87 & 12.83 & 17.39 & 49.42 & 12.71 & 19.43 & 42.57 & 10.50 & 14.21 & 36.07 & 10.14 & 14.71 & 39.71 & 20.61 & 27.32 & 60.86 \\
FID & 71.53 & 89.99 & 96.92 & 11.59 & 42.17 & 69.86 & 44.43 & 84.00 & 96.43 & 20.50 & 57.43 & 81.43 & 15.57 & 51.00 & 79.57 & 51.28 & 86.10 & 96.65 \\
\midrule
MINTIME & 11.44 & 55.22 & 91.13 & 6.45 & 47.17 & 85.80 & 1.14 & 8.57 & 31.14 & 9.57 & 52.36 & 86.07 & 5.29 & 26.86 & 58.71 & 7.35 & 38.50 & 80.03 \\
FTCN & 50.93 & 85.19 & 96.14 & 33.91 & 72.54 & 88.70 & 10.71 & 37.29 & 63.14 & 24.71 & 62.57 & 84.93 & 16.00 & 43.43 & 66.71 & 22.04 & 65.65 & 89.62 \\
TALL & 0.36 & 1.65 & 4.72 & 0.14 & 1.74 & 6.09 & 0.57 & 2.43 & 7.57 & 0.00 & 0.50 & 2.57 & 0.29 & 2.43 & 6.57 & 0.00 & 0.64 & 3.35 \\
XCLIP & 15.31 & 41.85 & 72.39 & 8.19 & 27.68 & 65.94 & 6.71 & 18.57 & 46.29 & 7.29 & 18.21 & 43.57 & 7.43 & 22.71 & 55.71 & 9.58 & 29.87 & 68.85 \\
AIGVDet & 67.38 & 87.84 & 95.06 & 50.22 & 71.74 & 84.71 & 0.00 & 0.00 & 0.00 & 9.29 & 21.21 & 33.57 & 38.43 & 65.29 & 82.29 & 79.07 & 94.09 & 97.60 \\
DeMamba & 15.81 & 45.42 & 72.68 & 16.30 & 49.06 & 78.55 & 2.57 & 15.43 & 38.14 & 2.29 & 16.21 & 42.07 & 2.43 & 14.71 & 37.71 & 1.76 & 11.98 & 31.63 \\
ReStraV & \textbf{96.49} & \textbf{99.71} & \textbf{100.00} & \textbf{96.52} & \textbf{99.64} & \textbf{100.00} & 99.43 & \textbf{100.00} & \textbf{100.00} & 92.07 & 97.43 & \textbf{99.00} & \textbf{94.43} & \textbf{97.71} & \textbf{99.14} & \textbf{89.78} & \textbf{99.84} & \textbf{100.00} \\
\midrule
D3 & 69.31 & 88.13 & 96.35 & 83.19 & 96.38 & 99.57 & 53.71 & 81.14 & 93.71 & 57.86 & 81.64 & 91.79 & 46.57 & 76.29 & 89.29 & 81.31 & 93.61 & 98.88 \\
\framework{} & 94.64 & 98.21 & 99.79 & 70.00 & 86.30 & 95.22 & \textbf{100.00} & \textbf{100.00} & \textbf{100.00} & \textbf{95.14} & \textbf{97.57} & 98.86 & 91.43 & 95.57 & 98.43 & 68.05 & 87.22 & 96.17 \\
\bottomrule
\end{tabular}%
}

\resizebox{\textwidth}{!}{%
\begin{tabular}{lcccccccccccccccc}
\toprule
\multirow{2}{*}{Method} & \multicolumn{3}{c}{MorphStudio} & \multicolumn{3}{c}{Show\_1} & \multicolumn{3}{c}{Sora} & \multicolumn{3}{c}{WildScrape} & & \multicolumn{3}{c}{Overall} \\
\cmidrule(lr){2-4} \cmidrule(lr){5-7} \cmidrule(lr){8-10} \cmidrule(lr){11-13} \cmidrule(lr){15-17}
 & 0.1\% & 1\% & 5\% & 0.1\% & 1\% & 5\% & 0.1\% & 1\% & 5\% & 0.1\% & 1\% & 5\% & & 0.1\% & 1\% & 5\% \\
\midrule
NPR & 0.14 & 1.43 & 10.71 & 0.00 & 0.00 & 0.43 & 0.00 & 0.00 & 10.71 & 0.11 & 1.75 & 9.74 & & 0.18 & 1.79 & 10.98 \\
STIL & 15.29 & 19.00 & 46.71 & 8.14 & 12.86 & 37.00 & 7.14 & 10.71 & 32.14 & 7.22 & 10.07 & 28.77 & & 12.73 & 17.40 & 43.01 \\
FID & 20.29 & 59.29 & 85.71 & 12.29 & 41.86 & 77.86 & 0.00 & 5.36 & 16.07 & 20.24 & 36.32 & 56.46 & & 26.77 & 55.35 & 75.70 \\
\midrule
MINTIME & 10.57 & 51.43 & 89.00 & 5.29 & 33.14 & 73.29 & 1.79 & 12.50 & 28.57 & 9.96 & 40.04 & 67.51 & & 6.88 & 36.58 & 69.12 \\
FTCN & 45.57 & 78.86 & 93.00 & 14.86 & 48.29 & 72.29 & 8.93 & 16.07 & 32.14 & 28.45 & 52.74 & 64.88 & & 25.61 & 56.26 & 75.16 \\
TALL & 0.71 & 2.29 & 6.86 & 0.00 & 2.14 & 6.57 & 0.00 & 0.00 & 1.79 & 0.11 & 0.66 & 4.05 & & 0.22 & 1.45 & 5.01 \\
XCLIP & 21.86 & 42.57 & 69.71 & 8.86 & 28.00 & 52.71 & 8.93 & 17.86 & 39.29 & 10.07 & 23.63 & 47.26 & & 10.42 & 27.09 & 56.17 \\
AIGVDet & 52.86 & 76.57 & 90.43 & 7.00 & 16.43 & 24.86 & 25.00 & 55.36 & 71.43 & 24.84 & 36.43 & 46.94 & & 35.41 & 52.50 & 62.69 \\
DeMamba & 13.14 & 38.43 & 66.14 & 7.14 & 29.43 & 58.43 & 1.79 & 8.93 & 25.00 & 5.91 & 20.68 & 40.70 & & 6.91 & 25.03 & 49.10 \\
ReStraV & 97.57 & \textbf{99.71} & \textbf{99.86} & 98.86 & \textbf{100.00} & \textbf{100.00} & \textbf{98.21} & \textbf{100.00} & \textbf{100.00} & \textbf{86.11} & \textbf{89.28} & \textbf{90.70} & & \textbf{94.95} & \textbf{98.33} & \textbf{98.87} \\
\midrule
D3 & 51.14 & 79.43 & 95.43 & 65.86 & 90.86 & 98.14 & 60.71 & 91.07 & 96.43 & 32.39 & 61.05 & 79.98 & & 60.20 & 83.96 & 93.96 \\
\framework{} & \textbf{97.71} & 99.29 & \textbf{99.86} & \textbf{99.43} & \textbf{100.00} & \textbf{100.00} & 94.64 & \textbf{100.00} & \textbf{100.00} & 77.13 & 82.06 & 86.65 & & 88.82 & 94.62 & 97.50 \\
\bottomrule
\end{tabular}%
}
\end{table}

\begin{table}[p]
\centering
\caption{Fake Recall (\%) on GenVideo at FPR $\in \{0.1\%,\,1\%,\,5\%\}$, with thresholds calibrated using the MSR-VTT real set only. }  
\label{tab:genvideo_msrvtt}
\renewcommand{\arraystretch}{1.0}
\setlength{\tabcolsep}{3pt}

\resizebox{\textwidth}{!}{%
\begin{tabular}{lcccccccccccccccccc}
\toprule
\multirow{2}{*}{Method}
 & \multicolumn{3}{c}{Crafter} & \multicolumn{3}{c}{Gen2} & \multicolumn{3}{c}{HotShot} & \multicolumn{3}{c}{Lavie} & \multicolumn{3}{c}{ModelScope} & \multicolumn{3}{c}{MoonValley} \\
\cmidrule(lr){2-4} \cmidrule(lr){5-7} \cmidrule(lr){8-10} \cmidrule(lr){11-13} \cmidrule(lr){14-16} \cmidrule(lr){17-19}
 & 0.1\% & 1\% & 5\% & 0.1\% & 1\% & 5\% & 0.1\% & 1\% & 5\% & 0.1\% & 1\% & 5\% & 0.1\% & 1\% & 5\% & 0.1\% & 1\% & 5\% \\
\midrule
NPR & 25.97 & 46.07 & 65.52 & 11.96 & 26.52 & 47.90 & 17.14 & 35.86 & 61.00 & 7.29 & 19.93 & 39.86 & 17.57 & 32.71 & 50.57 & 20.77 & 39.62 & 60.86 \\
STIL & 22.75 & 34.69 & 64.52 & 12.83 & 24.93 & 58.12 & 12.71 & 24.29 & 50.57 & 10.50 & 19.57 & 41.93 & 10.14 & 20.43 & 47.43 & 20.61 & 36.42 & 69.65 \\
FID & 87.48 & 97.85 & 99.57 & 35.43 & 77.46 & 93.91 & 77.43 & 97.29 & 99.43 & 50.07 & 87.14 & \textbf{98.86} & 43.14 & 86.86 & \textbf{97.86} & 82.43 & 98.72 & \textbf{99.84} \\
\midrule
MINTIME & 22.03 & 64.66 & 90.77 & 15.43 & 56.16 & 85.22 & 2.14 & 13.43 & 30.57 & 21.43 & 61.43 & 85.79 & 9.57 & 32.71 & 58.43 & 13.10 & 49.20 & 79.39 \\
FTCN & 76.32 & 91.06 & 97.07 & 60.43 & 82.25 & 91.67 & 28.43 & 50.14 & 71.14 & 51.14 & 76.36 & 88.43 & 34.00 & 56.00 & 72.57 & 51.12 & 78.27 & 92.97 \\
TALL & 0.64 & 3.58 & 11.52 & 0.51 & 4.35 & 17.25 & 1.57 & 5.00 & 15.29 & 0.00 & 1.71 & 5.71 & 0.71 & 5.14 & 14.00 & 0.16 & 2.24 & 8.95 \\
XCLIP & 9.87 & 43.63 & 75.75 & 4.71 & 28.99 & 71.38 & 4.71 & 19.43 & 51.43 & 4.79 & 19.29 & 47.43 & 5.00 & 24.29 & 60.86 & 6.55 & 31.31 & 72.84 \\
AIGVDet & \textbf{97.57} & \textbf{99.14} & 99.57 & \textbf{91.23} & \textbf{95.43} & 97.32 & 0.00 & 0.14 & 0.14 & 44.79 & 58.57 & 65.43 & \textbf{90.71} & \textbf{96.43} & \textbf{97.86} & \textbf{98.24} & \textbf{99.04} & 99.36 \\
DeMamba & 35.84 & 68.96 & 86.70 & 38.84 & 74.64 & 90.22 & 9.86 & 34.43 & 56.43 & 10.21 & 37.14 & 64.57 & 9.71 & 33.29 & 59.86 & 7.67 & 27.64 & 55.75 \\
ReStraV & 11.23 & 51.22 & 83.12 & 48.84 & 73.84 & 87.03 & 27.00 & 87.57 & 97.71 & 18.64 & 60.93 & 80.14 & 43.29 & 77.57 & 89.71 & 40.10 & 66.77 & 77.00 \\
\midrule
D3 & 8.44 & 66.38 & 91.77 & 4.42 & 80.80 & \textbf{97.90} & 1.57 & 49.14 & 85.43 & 1.14 & 54.29 & 85.14 & 1.00 & 40.71 & 80.43 & 11.02 & 78.43 & 95.53 \\
\framework{} & 88.84 & 96.78 & \textbf{99.64} & 53.26 & 77.61 & 94.71 & \textbf{99.86} & \textbf{100.00} & \textbf{100.00} & \textbf{92.50} & \textbf{96.36} & 98.79 & 86.71 & 93.57 & \textbf{97.86} & 49.36 & 76.36 & 95.37 \\
\bottomrule
\end{tabular}%
}

\resizebox{\textwidth}{!}{%
\begin{tabular}{lcccccccccccccccc}
\toprule
\multirow{2}{*}{Method} & \multicolumn{3}{c}{MorphStudio} & \multicolumn{3}{c}{Show\_1} & \multicolumn{3}{c}{Sora} & \multicolumn{3}{c}{WildScrape} & & \multicolumn{3}{c}{Overall} \\
\cmidrule(lr){2-4} \cmidrule(lr){5-7} \cmidrule(lr){8-10} \cmidrule(lr){11-13} \cmidrule(lr){15-17}
 & 0.1\% & 1\% & 5\% & 0.1\% & 1\% & 5\% & 0.1\% & 1\% & 5\% & 0.1\% & 1\% & 5\% & & 0.1\% & 1\% & 5\% \\
\midrule
NPR & 12.86 & 31.57 & 55.29 & 0.57 & 3.86 & 9.71 & 12.50 & 19.64 & 35.71 & 12.04 & 22.76 & 38.73 & & 13.87 & 27.85 & 46.52 \\
STIL & 15.29 & 25.57 & 53.57 & 8.14 & 16.29 & 43.71 & 7.14 & 14.29 & 33.93 & 7.22 & 13.57 & 33.59 & & 12.73 & 23.01 & 49.70 \\
FID & 51.43 & 90.57 & 98.71 & 35.00 & 86.29 & 98.43 & 1.79 & 25.00 & 58.93 & 33.26 & 63.02 & 79.43 & & 49.75 & 81.02 & 92.50 \\
\midrule
MINTIME & 20.86 & 60.29 & 88.86 & 10.29 & 40.71 & 72.71 & 3.57 & 16.07 & 28.57 & 18.71 & 47.59 & 67.07 & & 13.71 & 44.23 & 68.74 \\
FTCN & 70.00 & 86.86 & 96.29 & 34.71 & 60.43 & 79.57 & 12.50 & 23.21 & 33.93 & 47.37 & 58.86 & 70.35 & & 46.60 & 66.34 & 79.40 \\
TALL & 0.86 & 4.14 & 14.86 & 0.29 & 4.14 & 17.29 & 0.00 & 1.79 & 5.36 & 0.33 & 2.30 & 10.83 & & 0.51 & 3.44 & 12.11 \\
XCLIP & 16.86 & 44.00 & 73.57 & 5.43 & 29.57 & 55.86 & 3.57 & 17.86 & 48.21 & 7.11 & 24.62 & 50.66 & & 6.86 & 28.30 & 60.80 \\
AIGVDet & \textbf{96.71} & 98.29 & 99.14 & 33.43 & 46.71 & 54.57 & 76.79 & 87.50 & 89.29 & 57.22 & 65.75 & 68.71 & & 68.67 & 74.70 & 77.14 \\
DeMamba & 31.86 & 61.00 & 82.14 & 20.43 & 53.86 & 79.71 & 7.14 & 21.43 & 30.36 & 15.10 & 36.32 & 52.63 & & 18.67 & 44.87 & 65.84 \\
ReStraV & 23.71 & 69.29 & 88.29 & 25.14 & 78.29 & 93.86 & 48.21 & 80.36 & 92.86 & 25.27 & 61.49 & 78.01 & & 31.14 & 70.73 & 86.77 \\
\midrule
D3 & 1.43 & 47.71 & 85.14 & 0.71 & 61.00 & 93.43 & 0.00 & 50.00 & 92.86 & 0.22 & 28.99 & 66.52 & & 3.00 & 55.75 & 87.42 \\
\framework{} & 94.71 & \textbf{98.43} & \textbf{99.71} & \textbf{96.86} & \textbf{99.86} & \textbf{100.00} & \textbf{87.50} & \textbf{98.21} & \textbf{100.00} & \textbf{71.66} & \textbf{79.76} & \textbf{85.89} & & \textbf{82.13} & \textbf{91.69} & \textbf{97.20} \\
\bottomrule
\end{tabular}%
}
\end{table}
\begin{table}[t]
\centering
\caption{Fake Recall (\%) on ViF-Bench at FPR $= 0.1\%$, with thresholds calibrated using the ROVI real set only. Best results are in \textbf{bold}.}
\label{tab:vif_results_rovi_01}
\renewcommand{\arraystretch}{1.2}

\resizebox{\textwidth}{!}{%
\begin{tabular}{lcccccccccccccc|ccccc|c}
\toprule
\multirow{2}{*}{Method} & \multicolumn{14}{c}{T2V Models} & \multicolumn{5}{c}{I2V Models} & \multirow{2}{*}{Overall} \\
\cmidrule(lr){2-15} \cmidrule(lr){16-20}
 & \rotatebox{90}{CogX1.5} & \rotatebox{90}{Hunyuan} & \rotatebox{90}{LTX-T} & \rotatebox{90}{SkyV2} & \rotatebox{90}{Wan2.1} & \rotatebox{90}{Wan2.1V} & \rotatebox{90}{Wan2.2} & \rotatebox{90}{Wan2.2TI-T} & \rotatebox{90}{Gen4} & \rotatebox{90}{Hailuo-02} & \rotatebox{90}{Kling-V1} & \rotatebox{90}{Pika-V2} & \rotatebox{90}{Pix4.5} & \rotatebox{90}{Sora-2} & \rotatebox{90}{Hunyuan-I} & \rotatebox{90}{LTX-I} & \rotatebox{90}{SkyV2-I} & \rotatebox{90}{Wan2.2-I} & \rotatebox{90}{Wan2.2TI-I} &  \\
\midrule
NPR & 0.00 & 0.00 & 0.00 & 0.00 & 0.00 & 0.00 & 0.00 & 0.00 & 0.00 & 0.00 & 0.00 & 0.00 & 0.00 & 0.00 & 0.00 & 0.00 & 0.00 & 0.00 & 0.00 & 0.00 \\
STIL & 2.02 & 3.05 & 5.58 & 3.54 & 3.03 & 0.51 & 6.57 & 1.52 & 1.32 & 3.51 & 1.71 & 1.08 & 2.21 & 1.11 & 1.01 & 1.52 & 0.51 & 1.01 & 0.00 & 2.15 \\
FID & 6.57 & 3.55 & 0.51 & 3.54 & 7.07 & 0.51 & 2.53 & 0.51 & 7.89 & 3.51 & 0.57 & 15.14 & 2.76 & 0.56 & 0.00 & 3.54 & 1.52 & 5.56 & 0.51 & 3.49 \\
\midrule
MINTIME & 7.58 & 8.12 & 5.58 & 8.08 & 11.62 & 2.53 & 8.08 & 2.02 & 7.89 & 15.20 & 6.86 & 12.97 & 9.94 & 1.67 & 0.00 & 2.53 & 1.52 & 0.51 & 0.51 & 5.96 \\
FTCN & 13.64 & 18.78 & 8.63 & 17.17 & 11.62 & 4.55 & 11.62 & 1.01 & 12.50 & 35.09 & 16.00 & 21.62 & 30.94 & 2.22 & 0.00 & 6.06 & 2.54 & 3.03 & 1.52 & 11.50 \\
TALL & 0.00 & 0.51 & 0.00 & 0.00 & 0.00 & 0.00 & 0.00 & 0.00 & 0.00 & 0.00 & 0.00 & 0.00 & 0.00 & 0.00 & 0.00 & 0.00 & 0.00 & 0.00 & 0.51 & 0.05 \\
XCLIP & 1.52 & 4.06 & 0.00 & 5.05 & 0.00 & 0.51 & 3.03 & 1.01 & 0.00 & 5.26 & 0.57 & 0.54 & 0.55 & 1.11 & 0.00 & 1.01 & 0.00 & 0.00 & 0.00 & 1.27 \\
AIGVDet & 13.64 & 17.77 & 35.03 & 44.95 & 18.69 & 17.68 & 30.81 & 10.61 & 7.24 & 36.84 & 11.43 & 20.54 & 35.91 & 12.22 & 0.51 & 9.09 & 6.60 & 10.61 & 3.54 & 18.09 \\
DeMamba & 1.01 & 5.08 & 0.51 & 2.02 & 3.54 & 1.01 & 3.54 & 2.02 & 3.29 & 7.02 & 1.71 & 1.08 & 3.87 & 1.11 & 0.51 & 1.01 & 1.02 & 1.52 & 1.01 & 2.20 \\
ReStraV & \textbf{89.90} & \textbf{96.95} & \textbf{96.45} & \textbf{99.49} & \textbf{98.99} & \textbf{97.47} & \textbf{98.48} & 91.92 & \textbf{94.08} & \textbf{100.00} & \textbf{100.00} & \textbf{98.92} & \textbf{98.90} & \textbf{97.78} & \textbf{89.90} & \textbf{98.99} & \textbf{97.46} & \textbf{98.48} & 94.44 & \textbf{96.77} \\
\midrule
D3 & 42.93 & 31.98 & 73.60 & 47.47 & 36.87 & 27.78 & 46.46 & 29.80 & 31.58 & 37.43 & 63.43 & 29.19 & 54.14 & 36.11 & 35.35 & 31.31 & 38.58 & 30.30 & 19.70 & 39.16 \\
\framework{} & 87.37 & 88.32 & 79.70 & 86.87 & 95.45 & 95.96 & 86.36 & \textbf{98.99} & \textbf{94.08} & 78.95 & 97.71 & 98.38 & 83.43 & 96.67 & 70.71 & 88.89 & 89.34 & 94.44 & \textbf{100.00} & 90.09 \\
\bottomrule
\end{tabular}%
}
\end{table}

\begin{table}[t]
\centering
\caption{Fake Recall (\%) on ViF-Bench at FPR $= 1\%$, with thresholds calibrated using the ROVI real set only.}
\label{tab:vif_results_rovi_1}
\renewcommand{\arraystretch}{1.2}

\resizebox{\textwidth}{!}{%
\begin{tabular}{lcccccccccccccc|ccccc|c}
\toprule
\multirow{2}{*}{Method} & \multicolumn{14}{c}{T2V Models} & \multicolumn{5}{c}{I2V Models} & \multirow{2}{*}{Overall} \\
\cmidrule(lr){2-15} \cmidrule(lr){16-20}
 & \rotatebox{90}{CogX1.5} & \rotatebox{90}{Hunyuan} & \rotatebox{90}{LTX-T} & \rotatebox{90}{SkyV2} & \rotatebox{90}{Wan2.1} & \rotatebox{90}{Wan2.1V} & \rotatebox{90}{Wan2.2} & \rotatebox{90}{Wan2.2TI-T} & \rotatebox{90}{Gen4} & \rotatebox{90}{Hailuo-02} & \rotatebox{90}{Kling-V1} & \rotatebox{90}{Pika-V2} & \rotatebox{90}{Pix4.5} & \rotatebox{90}{Sora-2} & \rotatebox{90}{Hunyuan-I} & \rotatebox{90}{LTX-I} & \rotatebox{90}{SkyV2-I} & \rotatebox{90}{Wan2.2-I} & \rotatebox{90}{Wan2.2TI-I} &  \\
\midrule
NPR & 0.51 & 1.02 & 0.51 & 0.00 & 0.00 & 0.00 & 0.00 & 0.00 & 0.66 & 1.17 & 2.29 & 3.78 & 1.10 & 0.00 & 0.00 & 1.01 & 0.51 & 0.00 & 0.00 & 0.66 \\
STIL & 2.53 & 3.55 & 12.69 & 4.04 & 4.55 & 1.01 & 7.07 & 2.53 & 1.32 & 3.51 & 1.71 & 1.08 & 2.76 & 1.67 & 2.02 & 1.52 & 1.02 & 1.52 & 0.51 & 2.98 \\
FID & 30.81 & 19.80 & 9.64 & 26.26 & 20.71 & 11.11 & 11.62 & 4.04 & 15.79 & 23.98 & 11.43 & 45.95 & 17.13 & 3.89 & 4.04 & 9.60 & 7.11 & 12.12 & 7.07 & 15.37 \\
\midrule
MINTIME & 45.45 & 38.58 & 38.58 & 42.93 & 54.04 & 10.61 & 41.92 & 19.19 & 32.89 & 54.39 & 52.00 & 56.76 & 62.43 & 18.33 & 0.00 & 15.15 & 11.17 & 12.63 & 10.10 & 32.48 \\
FTCN & 37.37 & 40.61 & 49.24 & 47.98 & 36.87 & 14.14 & 40.40 & 9.09 & 27.63 & 63.16 & 44.00 & 54.59 & 61.88 & 9.44 & 0.51 & 15.15 & 13.20 & 15.15 & 8.59 & 31.00 \\
TALL & 1.52 & 0.51 & 0.51 & 1.01 & 1.01 & 1.52 & 0.51 & 0.00 & 1.97 & 0.00 & 0.00 & 0.00 & 0.55 & 0.56 & 0.51 & 2.53 & 0.51 & 1.52 & 0.51 & 0.80 \\
XCLIP & 4.55 & 12.18 & 2.03 & 16.16 & 6.57 & 4.04 & 13.64 & 3.54 & 5.26 & 14.04 & 2.29 & 3.78 & 6.63 & 3.89 & 0.51 & 4.55 & 2.54 & 4.04 & 0.51 & 5.83 \\
AIGVDet & 26.77 & 38.58 & 47.72 & 60.10 & 34.85 & 26.77 & 43.43 & 21.72 & 18.42 & 53.80 & 23.43 & 40.00 & 51.93 & 23.33 & 1.52 & 15.66 & 12.18 & 18.18 & 9.09 & 29.87 \\
DeMamba & 9.09 & 17.77 & 3.55 & 21.21 & 15.15 & 5.56 & 18.18 & 7.07 & 11.18 & 29.24 & 11.43 & 6.49 & 19.89 & 8.89 & 1.01 & 6.06 & 6.09 & 6.57 & 3.54 & 10.95 \\
ReStraV & \textbf{98.99} & \textbf{98.98} & \textbf{100.00} & \textbf{100.00} & \textbf{100.00} & \textbf{99.49} & \textbf{100.00} & 98.48 & \textbf{99.34} & \textbf{100.00} & \textbf{100.00} & \textbf{100.00} & \textbf{99.45} & \textbf{99.44} & \textbf{96.46} & \textbf{100.00} & \textbf{100.00} & \textbf{99.49} & \textbf{100.00} & \textbf{99.48} \\
\midrule
D3 & 71.21 & 75.13 & 92.89 & 87.37 & 74.75 & 68.18 & 83.33 & 61.62 & 73.68 & 80.12 & 90.29 & 69.73 & 85.64 & 72.22 & 73.23 & 74.75 & 78.68 & 74.24 & 63.13 & 76.33 \\
\framework{} & 90.40 & 95.43 & 91.88 & 97.98 & 99.49 & \textbf{99.49} & 96.97 & \textbf{100.00} & 97.37 & 92.40 & \textbf{100.00} & 99.46 & 98.34 & 98.89 & 86.87 & 95.96 & 96.95 & 97.47 & \textbf{100.00} & 96.60 \\
\bottomrule
\end{tabular}%
}
\end{table}

\begin{table}[t]
\centering
\caption{Fake Recall (\%) on ViF-Bench at FPR $= 5\%$, with thresholds calibrated using the ROVI real set only.}
\label{tab:vif_results_rovi_5}
\renewcommand{\arraystretch}{1.2}

\resizebox{\textwidth}{!}{%
\begin{tabular}{lcccccccccccccc|ccccc|c}
\toprule
\multirow{2}{*}{Method} & \multicolumn{14}{c}{T2V Models} & \multicolumn{5}{c}{I2V Models} & \multirow{2}{*}{Overall} \\
\cmidrule(lr){2-15} \cmidrule(lr){16-20}
 & \rotatebox{90}{CogX1.5} & \rotatebox{90}{Hunyuan} & \rotatebox{90}{LTX-T} & \rotatebox{90}{SkyV2} & \rotatebox{90}{Wan2.1} & \rotatebox{90}{Wan2.1V} & \rotatebox{90}{Wan2.2} & \rotatebox{90}{Wan2.2TI-T} & \rotatebox{90}{Gen4} & \rotatebox{90}{Hailuo-02} & \rotatebox{90}{Kling-V1} & \rotatebox{90}{Pika-V2} & \rotatebox{90}{Pix4.5} & \rotatebox{90}{Sora-2} & \rotatebox{90}{Hunyuan-I} & \rotatebox{90}{LTX-I} & \rotatebox{90}{SkyV2-I} & \rotatebox{90}{Wan2.2-I} & \rotatebox{90}{Wan2.2TI-I} &  \\
\midrule
NPR & 4.04 & 9.64 & 1.02 & 5.56 & 5.05 & 0.51 & 0.00 & 0.00 & 10.53 & 9.94 & 10.29 & 10.27 & 5.52 & 0.00 & 0.51 & 3.54 & 3.55 & 0.51 & 1.01 & 4.29 \\
STIL & 10.10 & 12.69 & 30.46 & 16.16 & 15.66 & 5.56 & 14.14 & 9.09 & 4.61 & 9.94 & 10.29 & 8.11 & 11.60 & 7.78 & 7.07 & 4.55 & 3.55 & 5.56 & 2.53 & 9.97 \\
FID & 47.47 & 40.10 & 27.92 & 44.44 & 43.43 & 18.18 & 30.30 & 10.10 & 21.05 & 45.61 & 26.86 & 72.43 & 38.12 & 10.56 & 9.09 & 16.16 & 15.74 & 16.16 & 11.11 & 28.68 \\
\midrule
MINTIME & 65.66 & 70.05 & 69.04 & 71.72 & 85.35 & 37.88 & 71.21 & 46.97 & 66.45 & 84.21 & 84.57 & 84.86 & 91.16 & 42.78 & 2.02 & 42.42 & 33.50 & 33.84 & 25.76 & 58.39 \\
FTCN & 59.60 & 60.91 & 67.01 & 65.66 & 66.16 & 29.80 & 61.11 & 22.73 & 43.42 & 77.78 & 64.57 & 78.92 & 78.45 & 27.22 & 2.02 & 25.76 & 21.83 & 21.72 & 14.65 & 46.81 \\
TALL & 2.02 & 5.08 & 3.05 & 2.53 & 2.53 & 7.07 & 5.56 & 1.01 & 7.89 & 2.34 & 1.71 & 0.54 & 2.76 & 2.22 & 10.10 & 8.59 & 4.57 & 8.59 & 6.57 & 4.46 \\
XCLIP & 17.17 & 31.98 & 19.29 & 36.87 & 20.71 & 11.11 & 34.85 & 10.61 & 14.47 & 34.50 & 18.29 & 12.43 & 29.28 & 12.78 & 2.02 & 10.10 & 9.14 & 11.62 & 6.57 & 18.09 \\
AIGVDet & 37.88 & 52.28 & 62.94 & 69.19 & 52.02 & 32.83 & 55.05 & 28.79 & 23.03 & 72.51 & 34.29 & 52.43 & 61.33 & 28.89 & 2.02 & 21.21 & 19.80 & 22.73 & 15.15 & 39.18 \\
DeMamba & 29.29 & 39.09 & 28.43 & 48.99 & 38.89 & 18.18 & 38.89 & 23.23 & 23.68 & 57.31 & 32.00 & 24.32 & 52.49 & 25.00 & 6.57 & 17.17 & 15.74 & 17.68 & 15.66 & 29.08 \\
ReStraV & \textbf{100.00} & \textbf{100.00} & \textbf{100.00} & \textbf{100.00} & \textbf{100.00} & \textbf{100.00} & \textbf{100.00} & \textbf{100.00} & \textbf{99.34} & \textbf{100.00} & \textbf{100.00} & \textbf{100.00} & \textbf{100.00} & \textbf{100.00} & \textbf{98.48} & \textbf{100.00} & \textbf{100.00} & \textbf{100.00} & \textbf{100.00} & \textbf{99.89} \\
\midrule
D3 & 92.93 & 92.89 & \textbf{100.00} & 98.99 & 92.42 & 90.40 & 96.97 & 90.40 & 92.11 & 97.66 & 99.43 & 92.97 & 95.58 & 95.56 & 91.92 & 95.45 & 93.91 & 94.95 & 88.89 & 94.39 \\
\framework{} & 92.42 & 97.97 & 96.95 & 98.99 & 99.49 & 99.49 & \textbf{100.00} & \textbf{100.00} & 98.68 & 97.66 & \textbf{100.00} & \textbf{100.00} & \textbf{100.00} & 99.44 & 95.96 & 96.97 & 98.98 & 98.48 & \textbf{100.00} & 98.50 \\
\bottomrule
\end{tabular}%
}
\end{table}
\begin{table}[t]
\centering
\caption{Fake Recall (\%) on ViF-Bench at FPR $= 0.1\%$, with thresholds calibrated using the MSR-VTT real set only. Best results are in \textbf{bold}.}
\label{tab:vif_results_msrvtt_01}
\renewcommand{\arraystretch}{1.2}
\resizebox{\textwidth}{!}{%
\begin{tabular}{lcccccccccccccc|ccccc|c}
\toprule
\multirow{2}{*}{Method} & \multicolumn{14}{c}{T2V Models} & \multicolumn{5}{c}{I2V Models} & \multirow{2}{*}{Overall} \\
\cmidrule(lr){2-15} \cmidrule(lr){16-20}
 & \rotatebox{90}{CogX1.5} & \rotatebox{90}{Hunyuan} & \rotatebox{90}{LTX-T} & \rotatebox{90}{SkyV2} & \rotatebox{90}{Wan2.1} & \rotatebox{90}{Wan2.1V} & \rotatebox{90}{Wan2.2} & \rotatebox{90}{Wan2.2TI-T} & \rotatebox{90}{Gen4} & \rotatebox{90}{Hailuo-02} & \rotatebox{90}{Kling-V1} & \rotatebox{90}{Pika-V2} & \rotatebox{90}{Pix4.5} & \rotatebox{90}{Sora-2} & \rotatebox{90}{Hunyuan-I} & \rotatebox{90}{LTX-I} & \rotatebox{90}{SkyV2-I} & \rotatebox{90}{Wan2.2-I} & \rotatebox{90}{Wan2.2TI-I} &  \\
\midrule
NPR & 6.57 & 12.18 & 2.03 & 8.08 & 6.57 & 1.01 & 0.00 & 0.00 & 13.16 & 12.28 & 13.14 & 14.05 & 7.73 & 0.00 & 0.51 & 7.58 & 5.08 & 0.51 & 1.01 & 5.87 \\
STIL & 2.02 & 3.05 & 5.58 & 3.54 & 3.03 & 0.51 & 6.57 & 1.52 & 1.32 & 3.51 & 1.71 & 1.08 & 2.21 & 1.11 & 1.01 & 1.52 & 0.51 & 1.01 & 0.00 & 2.15 \\
FID & 27.27 & 17.26 & 7.61 & 21.72 & 16.16 & 9.09 & 8.59 & 4.04 & 14.47 & 18.13 & 8.57 & 37.30 & 14.36 & 2.78 & 3.54 & 9.09 & 5.58 & 11.11 & 5.05 & 12.72 \\
\midrule
MINTIME & 20.71 & 15.74 & 11.17 & 15.66 & 22.73 & 3.54 & 16.67 & 5.56 & 11.18 & 27.49 & 14.86 & 28.65 & 25.41 & 7.22 & 0.00 & 6.57 & 3.55 & 2.02 & 2.53 & 12.70 \\
FTCN & 29.29 & 35.53 & 38.07 & 36.87 & 27.78 & 9.60 & 29.80 & 5.05 & 25.66 & 54.97 & 36.57 & 41.62 & 53.59 & 7.78 & 0.00 & 11.11 & 9.64 & 10.61 & 7.07 & 24.77 \\
TALL & 0.00 & 0.51 & 0.00 & 0.51 & 0.00 & 0.00 & 0.00 & 0.00 & 0.00 & 0.00 & 0.00 & 0.00 & 0.00 & 0.00 & 0.00 & 0.00 & 0.00 & 0.51 & 0.51 & 0.11 \\
XCLIP & 1.01 & 2.03 & 0.00 & 2.53 & 0.00 & 0.51 & 2.02 & 0.00 & 0.00 & 1.75 & 0.57 & 0.00 & 0.55 & 0.00 & 0.00 & 1.01 & 0.00 & 0.00 & 0.00 & 0.63 \\
AIGVDet & 45.96 & 63.45 & \textbf{70.56} & \textbf{78.79} & 60.10 & 40.91 & 66.67 & 35.35 & 27.63 & \textbf{79.53} & 45.71 & 65.95 & \textbf{70.17} & 39.44 & 4.55 & 24.75 & 24.37 & 26.77 & 22.73 & 47.02 \\
DeMamba & 6.06 & 13.71 & 1.52 & 13.64 & 11.11 & 2.53 & 10.61 & 4.04 & 9.21 & 16.96 & 8.57 & 3.24 & 12.15 & 3.33 & 1.01 & 4.04 & 3.05 & 5.05 & 2.53 & 6.97 \\
ReStraV & 9.60 & 22.34 & 13.20 & 27.27 & 22.73 & 16.67 & 21.72 & 11.62 & 14.47 & 31.58 & 34.86 & 29.73 & 20.99 & 11.11 & 19.70 & 20.71 & 31.98 & 30.81 & 9.09 & 21.06 \\
\midrule
D3 & 8.59 & 0.00 & 1.02 & 0.00 & 0.00 & 0.00 & 0.00 & 0.00 & 0.00 & 0.00 & 0.57 & 0.00 & 0.00 & 0.00 & 0.51 & 0.00 & 0.00 & 0.00 & 0.00 & 0.56 \\
\framework{} & \textbf{82.32} & \textbf{74.11} & 69.04 & 64.14 & \textbf{88.89} & \textbf{90.91} & \textbf{75.76} & \textbf{94.44} & \textbf{81.58} & 63.74 & \textbf{89.14} & \textbf{95.68} & 62.43 & \textbf{87.22} & \textbf{54.55} & \textbf{76.77} & \textbf{77.66} & \textbf{90.91} & \textbf{97.98} & \textbf{79.86} \\
\bottomrule
\end{tabular}%
}
\end{table}

\begin{table}[t]
\centering
\caption{Fake Recall (\%) on ViF-Bench at FPR $= 1\%$, with thresholds calibrated using the MSR-VTT real set only.}
\label{tab:vif_results_msrvtt_1}
\renewcommand{\arraystretch}{1.2}
\resizebox{\textwidth}{!}{%
\begin{tabular}{lcccccccccccccc|ccccc|c}
\toprule
\multirow{2}{*}{Method} & \multicolumn{14}{c}{T2V Models} & \multicolumn{5}{c}{I2V Models} & \multirow{2}{*}{Overall} \\
\cmidrule(lr){2-15} \cmidrule(lr){16-20}
 & \rotatebox{90}{CogX1.5} & \rotatebox{90}{Hunyuan} & \rotatebox{90}{LTX-T} & \rotatebox{90}{SkyV2} & \rotatebox{90}{Wan2.1} & \rotatebox{90}{Wan2.1V} & \rotatebox{90}{Wan2.2} & \rotatebox{90}{Wan2.2TI-T} & \rotatebox{90}{Gen4} & \rotatebox{90}{Hailuo-02} & \rotatebox{90}{Kling-V1} & \rotatebox{90}{Pika-V2} & \rotatebox{90}{Pix4.5} & \rotatebox{90}{Sora-2} & \rotatebox{90}{Hunyuan-I} & \rotatebox{90}{LTX-I} & \rotatebox{90}{SkyV2-I} & \rotatebox{90}{Wan2.2-I} & \rotatebox{90}{Wan2.2TI-I} &  \\
\midrule
NPR & 15.15 & 24.87 & 10.66 & 18.18 & 14.65 & 2.53 & 2.53 & 0.00 & 22.37 & 25.15 & 29.71 & 29.19 & 18.78 & 1.67 & 1.01 & 14.14 & 10.15 & 4.04 & 1.52 & 12.96 \\
STIL & 3.54 & 4.06 & 20.81 & 7.58 & 6.06 & 1.01 & 7.07 & 4.55 & 1.32 & 5.26 & 2.86 & 1.62 & 4.97 & 1.67 & 4.04 & 2.53 & 2.03 & 2.53 & 1.01 & 4.45 \\
FID & 54.04 & 45.18 & 34.52 & 51.52 & 49.49 & 21.72 & 36.36 & 12.12 & 23.68 & 47.95 & 32.57 & 77.84 & 45.30 & 16.67 & 14.65 & 19.19 & 18.78 & 18.18 & 11.62 & 33.23 \\
\midrule
MINTIME & 50.51 & 46.70 & 48.22 & 52.53 & 61.11 & 15.15 & 48.48 & 21.72 & 38.82 & 63.16 & 58.29 & 62.16 & 69.61 & 23.33 & 0.00 & 18.18 & 14.72 & 14.14 & 12.63 & 37.87 \\
FTCN & 50.00 & 56.35 & 60.91 & 60.61 & 51.01 & 19.70 & 53.03 & 16.67 & 34.87 & 71.35 & 54.29 & 69.19 & 71.82 & 18.89 & 0.51 & 20.20 & 16.75 & 17.68 & 11.62 & 39.76 \\
TALL & 2.02 & 3.55 & 1.02 & 1.01 & 1.52 & 3.54 & 3.54 & 0.00 & 5.92 & 0.58 & 0.57 & 0.54 & 0.55 & 1.67 & 7.07 & 6.57 & 2.03 & 7.07 & 5.05 & 2.83 \\
XCLIP & 5.05 & 13.20 & 2.54 & 17.68 & 7.07 & 4.04 & 14.65 & 3.54 & 7.24 & 14.62 & 2.29 & 3.78 & 7.73 & 3.89 & 0.51 & 4.55 & 3.05 & 4.04 & 0.51 & 6.31 \\
AIGVDet & 56.06 & 73.10 & 75.13 & 85.86 & 67.68 & 57.58 & 75.25 & 47.47 & 37.50 & \textbf{87.72} & 61.14 & 76.22 & 81.77 & 52.78 & 10.10 & 36.36 & 36.55 & 37.88 & 29.80 & 57.16 \\
DeMamba & 24.75 & 36.55 & 20.30 & 45.96 & 33.33 & 14.65 & 33.33 & 19.19 & 21.71 & 52.63 & 29.14 & 18.92 & 47.51 & 22.22 & 5.05 & 13.13 & 12.69 & 16.16 & 14.14 & 25.33 \\
ReStraV & 51.01 & 69.54 & 57.87 & 69.19 & 74.75 & 60.10 & 70.71 & 53.03 & 57.24 & 70.18 & 82.29 & 73.51 & 61.33 & 58.33 & 54.55 & 60.10 & 71.07 & 73.74 & 47.47 & 64.00 \\
\midrule
D3 & 39.39 & 26.90 & 68.02 & 43.43 & 30.81 & 23.23 & 39.90 & 23.74 & 28.29 & 34.50 & 57.71 & 23.24 & 49.17 & 33.89 & 31.82 & 25.25 & 31.47 & 26.26 & 16.67 & 34.40 \\
\framework{} & \textbf{89.39} & \textbf{92.89} & \textbf{85.79} & \textbf{92.93} & \textbf{96.97} & \textbf{98.48} & \textbf{91.92} & \textbf{99.49} & \textbf{94.74} & 87.13 & \textbf{99.43} & \textbf{99.46} & \textbf{91.16} & \textbf{97.22} & \textbf{77.78} & \textbf{93.43} & \textbf{93.91} & \textbf{96.97} & \textbf{100.00} & \textbf{93.64} \\
\bottomrule
\end{tabular}%
}
\end{table}

\begin{table}[t]
\centering
\caption{Fake Recall (\%) on ViF-Bench at FPR $= 5\%$, with thresholds calibrated using the MSR-VTT real set only.}
\label{tab:vif_results_msrvtt_5}
\renewcommand{\arraystretch}{1.2}
\resizebox{\textwidth}{!}{%
\begin{tabular}{lcccccccccccccc|ccccc|c}
\toprule
\multirow{2}{*}{Method} & \multicolumn{14}{c}{T2V Models} & \multicolumn{5}{c}{I2V Models} & \multirow{2}{*}{Overall} \\
\cmidrule(lr){2-15} \cmidrule(lr){16-20}
 & \rotatebox{90}{CogX1.5} & \rotatebox{90}{Hunyuan} & \rotatebox{90}{LTX-T} & \rotatebox{90}{SkyV2} & \rotatebox{90}{Wan2.1} & \rotatebox{90}{Wan2.1V} & \rotatebox{90}{Wan2.2} & \rotatebox{90}{Wan2.2TI-T} & \rotatebox{90}{Gen4} & \rotatebox{90}{Hailuo-02} & \rotatebox{90}{Kling-V1} & \rotatebox{90}{Pika-V2} & \rotatebox{90}{Pix4.5} & \rotatebox{90}{Sora-2} & \rotatebox{90}{Hunyuan-I} & \rotatebox{90}{LTX-I} & \rotatebox{90}{SkyV2-I} & \rotatebox{90}{Wan2.2-I} & \rotatebox{90}{Wan2.2TI-I} &  \\
\midrule
NPR & 38.89 & 40.10 & 31.98 & 31.82 & 30.30 & 9.09 & 5.05 & 5.05 & 31.58 & 39.77 & 46.86 & 52.97 & 35.36 & 8.33 & 3.03 & 24.24 & 22.34 & 11.62 & 3.03 & 24.81 \\
STIL & 14.65 & 13.20 & 34.01 & 20.71 & 21.72 & 7.07 & 16.16 & 12.12 & 5.26 & 11.70 & 12.00 & 11.89 & 16.57 & 10.56 & 8.08 & 5.56 & 4.57 & 6.57 & 4.55 & 12.47 \\
FID & 76.77 & 71.57 & 73.10 & 75.25 & 74.75 & 34.34 & 69.19 & 30.30 & 38.16 & 79.53 & 62.29 & 92.43 & 72.93 & 43.33 & 40.91 & 36.36 & 29.95 & 28.28 & 22.73 & 55.38 \\
\midrule
MINTIME & 65.66 & 69.54 & 69.04 & 71.21 & 85.35 & 37.37 & 70.71 & 46.97 & 66.45 & 83.04 & 84.57 & 84.86 & 90.61 & 42.22 & 2.02 & 40.91 & 32.99 & 33.84 & 25.76 & 58.06 \\
FTCN & 73.23 & 68.02 & 71.57 & 71.72 & 68.18 & 34.34 & 65.15 & 28.28 & 50.00 & 84.21 & 68.57 & 82.70 & 81.77 & 33.89 & 3.03 & 30.30 & 25.89 & 25.25 & 18.69 & 51.83 \\
TALL & 4.55 & 8.12 & 19.80 & 8.59 & 9.09 & 13.64 & 10.10 & 3.54 & 16.45 & 8.19 & 5.71 & 5.95 & 12.15 & 6.67 & 18.18 & 17.17 & 11.17 & 14.65 & 12.12 & 10.83 \\
XCLIP & 20.71 & 35.03 & 26.40 & 41.92 & 22.73 & 14.65 & 37.88 & 13.64 & 17.11 & 38.01 & 21.71 & 14.59 & 35.36 & 17.22 & 2.53 & 12.63 & 10.66 & 12.63 & 7.58 & 21.21 \\
AIGVDet & 63.13 & 78.17 & 81.22 & 90.40 & 75.25 & 63.64 & 80.81 & 53.54 & 50.66 & 91.23 & 70.29 & 82.70 & 86.74 & 61.67 & 15.15 & 48.48 & 43.15 & 46.97 & 34.85 & 64.11 \\
DeMamba & 46.46 & 60.41 & 58.88 & 66.67 & 58.59 & 34.85 & 57.07 & 39.90 & 34.87 & 74.27 & 54.86 & 45.95 & 70.17 & 43.33 & 11.62 & 32.83 & 26.90 & 28.28 & 26.77 & 45.93 \\
ReStraV & 80.30 & 90.86 & 81.73 & 89.90 & 95.45 & 90.91 & 90.91 & 82.32 & 83.55 & 91.81 & 97.71 & 94.05 & 86.74 & 87.22 & 83.33 & 89.90 & 88.83 & 91.41 & 75.25 & 88.01 \\
\midrule
D3 & 76.77 & 81.73 & 94.92 & 92.42 & 80.81 & 74.75 & 87.37 & 66.67 & 80.26 & 85.96 & 93.71 & 76.22 & 89.50 & 79.44 & 81.31 & 83.84 & 84.77 & 82.83 & 68.18 & 82.18 \\
\framework{} & \textbf{91.41} & \textbf{97.97} & \textbf{96.95} & \textbf{98.99} & \textbf{99.49} & \textbf{99.49} & \textbf{100.00} & \textbf{100.00} & \textbf{98.68} & \textbf{97.66} & \textbf{100.00} & \textbf{100.00} & \textbf{100.00} & \textbf{99.44} & \textbf{95.45} & \textbf{96.97} & \textbf{98.98} & \textbf{98.48} & \textbf{100.00} & \textbf{98.42} \\
\bottomrule
\end{tabular}%
}
\end{table}

\end{document}